\documentclass[10pt,twocolumn,letterpaper]{article}
\usepackage{wacv}
\usepackage{graphicx}
\usepackage{amsmath}
\usepackage{amssymb}
\usepackage{booktabs}
\usepackage{anyfontsize}
\usepackage[pagebackref,breaklinks,colorlinks]{hyperref}
\usepackage[capitalize]{cleveref}
\crefname{section}{Sec.}{Secs.}
\Crefname{section}{Section}{Sections}
\Crefname{table}{Table}{Tables}
\crefname{table}{Tab.}{Tabs.}

\usepackage{epsfig}
\usepackage{multirow}
\usepackage{makecell}
\usepackage{comment}
\usepackage{cellspace}
\usepackage{amsfonts}
\usepackage{color}
\usepackage{enumitem}
\usepackage{array}
\usepackage{algorithm}
\usepackage{algpseudocode}
\algrenewcommand\algorithmicrequire{\textbf{Inputs:}}
\algrenewcommand\algorithmicensure{\textbf{Outputs:}}
\usepackage[symbol]{footmisc}
\usepackage[accsupp]{axessibility}
\algrenewcommand\algorithmicindent{0.75em}

\def\eqref#1{(\ref{eq:#1})}
\def\eqlabel#1{\label{eq:#1}}
\def\figref#1{\ref{fig:#1}}
\def\figlabel#1{\label{fig:#1}}

\def\eqref#1{(\ref{eq:#1})}
\def\eqlabel#1{\label{eq:#1}}
\def\figref#1{\ref{fig:#1}}
\def\figlabel#1{\label{fig:#1}}

\def\xcomment#1{\textcolor[rgb]{.3,.3,.1}{\text{$/\!\!/$ {\em #1}}}}
\def\comment#1{\kern-1cm\xcomment{#1}}
\def\eqcomment#1{\kern-1cm\xcomment{#1}}

\def\m#1{\ensuremath{\mathtt{#1}}}

\def\v#1{\ensuremath{\mathbf{#1}}}

\def\real{\mathbb{R}}

\def\norm#1{\left\lVert#1\right\rVert}

\def\l2#1{\norm{#1}_2}

\begin{document}

\title{FUN-AD: Fully Unsupervised Learning for Anomaly Detection with Noisy Training Data}
\hypersetup{linkcolor=black}
\author{Jiin Im$^{1*}$ \quad Yongho Son$^{2*}$ \quad Je Hyeong Hong$^{1,2}$\footnotemark[2]\\
$^1$Dept. Electronic Engineering, Hanyang University \quad $^2$Dept. Artificial Intelligence, Hanyang University}
\maketitle

\begin{abstract}
While the mainstream research in anomaly detection has mainly followed the one-class classification, practical industrial environments often incur noisy training data due to annotation errors or lack of labels for new or refurbished products.
To address these issues, we propose a novel learning-based approach for fully unsupervised anomaly detection with unlabeled and potentially contaminated training data.
Our method is motivated by two observations, that i) the pairwise feature distances between the normal samples are on average likely to be smaller than those between the anomaly samples or heterogeneous samples and ii) pairs of features mutually closest to each other are likely to be homogeneous pairs, which hold if the normal data has smaller variance than the anomaly data.
Building on the first observation that nearest-neighbor distances can distinguish between confident normal samples and anomalies, we propose a pseudo-labeling strategy using an iteratively reconstructed memory bank (IRMB).
The second observation is utilized as a new loss function to promote class-homogeneity between mutually closest pairs thereby reducing the ill-posedness of the task.
Experimental results on two public industrial anomaly benchmarks and semantic anomaly examples validate the effectiveness of FUN-AD across different scenarios and anomaly-to-normal ratios. 
Our code is available at \href{https://github.com/HY-Vision-Lab/FUNAD}{https://github.com/HY-Vision-Lab/FUNAD}.
\end{abstract}
\footnotetext[1]{indicates equal contributions.}
\footnotetext[2]{Corresponding author}

\section{Introduction}
\label{sec:intro}

\begin{figure}[t]
\vspace{-2mm}
\centering
\includegraphics[width=0.96\columnwidth]{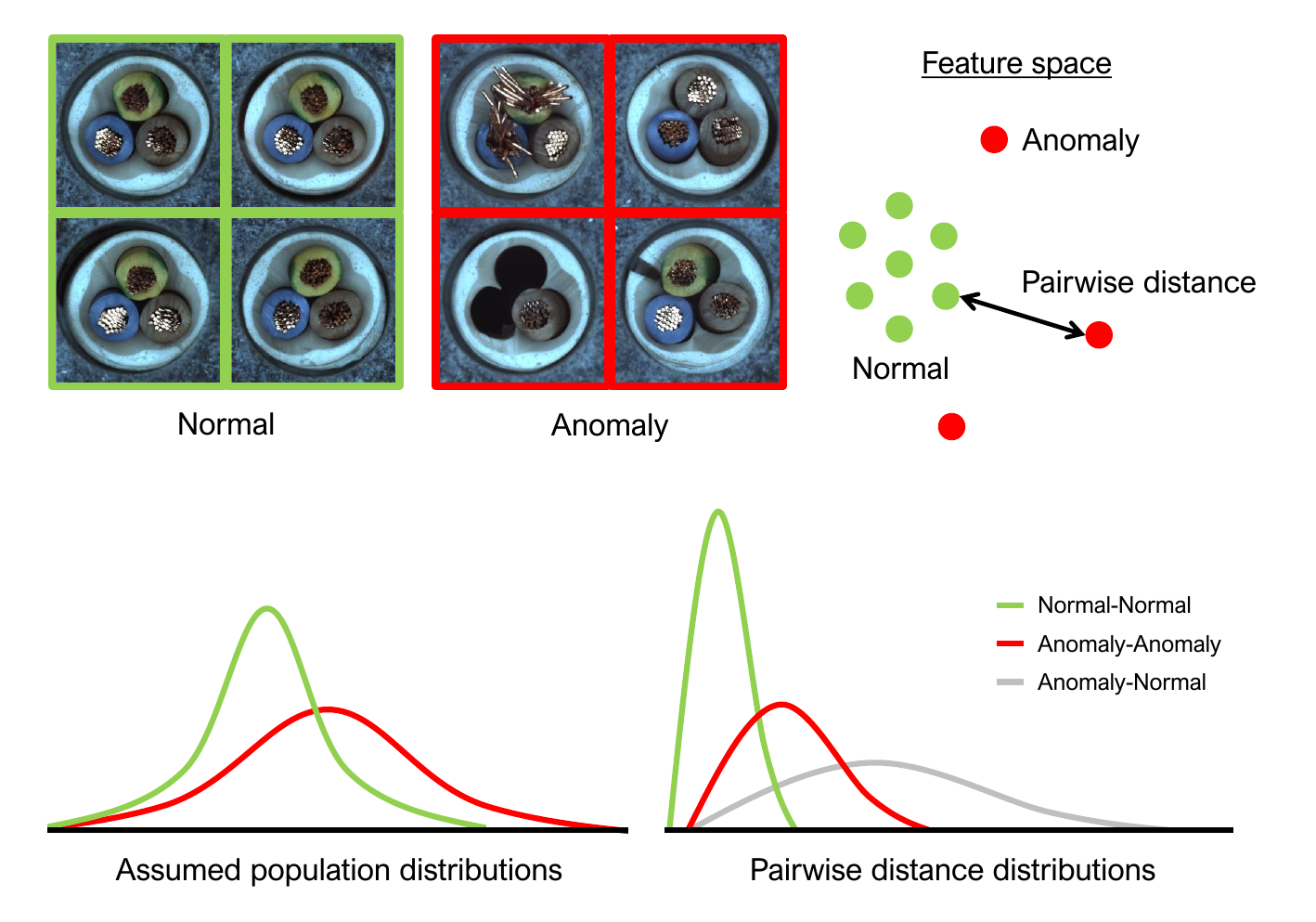}
\caption{
    An illustration of our motivation.
    We assume that the (image-level and patch-level) features from the normal samples (in green) exhibit smaller variance than those from the anomaly samples (in red).
    In Sec.~\ref{sec:statistical_analysis_pairwise_dist}, we analytically show this leads to a homogeneous pair of normal samples being more likely to yield smaller pairwise distance than other types of pairs.
    In Sec.~\ref{sec:smoothness_motivation}, we empirically show that mutually closest pairs are likely to be homogeneous pairs (i.e. mostly normal-normal or anomaly-anomaly).
}
\vspace{-5mm}
\label{fig:motivation}
\end{figure}

Anomaly detection refers to the process of detecting events that are rare and interesting, and is an essential application in engineering, science, medicines and finance~\cite{ma2021comprehensive, xie2023iad, chalapathy2019deep}.
In particular, anomaly detection involving visual data has received much attention recently, ranging from defect detection in manufacturing industry 
~\cite{deng2022anomaly, roth2022towards, wang2021glancing, zhang2023destseg, liu2023simplenet, defard2021padim, rudolph2022fully, tien2023revisiting, cohen2020sub, yao2023explicit, zhang2023prototypical, ding2022catching, shin2023anomaly, zavrtanik2021draem, li2021cutpaste, schluter2022natural, bae2023pni, zou2022spot, xi2022softpatch, zavrtanik2022dsr, mcintosh2023inter}, lesion detection in medical imaging~\cite{wolleb2022diffusion, xiang2023squid}
to violence detection in surveillance~\cite{zaheer2022generative, feng2021mist, pang2020self, yu2022deep, al2023coarse}.

While the task of industrial anomaly detection is relatively well-defined, there are two main issues that raise the difficulty of the problem in practice.
First, anomaly samples are rare and consequently difficult to obtain, triggering significant data imbalance between the normal and abnormal classes. 
Second, anomalies can arise from different causes, leading to a diverse distribution of anomaly samples.
Due to these problems, it is a commonly adopted problem setting in industrial anomaly detection~\cite{deng2022anomaly, roth2022towards, wang2021glancing, zhang2023destseg, liu2023simplenet, defard2021padim, rudolph2022fully, tien2023revisiting, shin2023anomaly, zavrtanik2021draem, li2021cutpaste, schluter2022natural, bae2023pni, zavrtanik2022dsr, cohen2020sub} that only the class of normal data is used for training.
While these methods are often noted as ``unsupervised'' approaches, they mostly adopt a form of supervised learning called one-class training since the training data requires correctly labeled normal samples. 
When training data is contaminated, one-class classification methods which do not separately consider outliers in the training data (e.g.~OC-SVM~\cite{manevitz2001one} does consider outliers) are vulnerable to contamination and may continue to mistake certain classes of anomalies for normals as they consider anomalies in the training dataset to be normals.
Consequently, this problem setting still requires clean data collection, which incurs considerable annotation costs and time. 
In this study, we explore the possibility of addressing the challenging  question ``can we train an accurate industrial anomaly detection algorithm without any labeled data?''.
In real-world scenarios, normal data can easily become outdated due to regular product upgrades or changes in manufacturing processes.
Moreover, even labeled training data can be contaminated with anomalies due to human errors in annotation.
All these issues with one-class training would be eradicated in the fully unsupervised setting, providing practical motivation.

While fully unsupervised approaches~\cite{xi2022softpatch, li2021deep, yoon2022selfsupervise, mcintosh2023inter, chen2022deep} exist to address the above questions, they mostly focus on eliminating samples pseudo-labeled as anomalies and performing one-class training with the remaining data. 
However, supervised anomaly detection studies~\cite{zhang2023prototypical, yao2023explicit} have shown that even limited anomaly information can enhance both anomaly detection and localization performance, while also addressing the issue of small anomalies being overlooked by traditional one-class classification methods.

To this end, we take a different approach towards using anomaly information compared to other fully unsupervised schemes.
Our work leverages statistical observations that i) pairs of normal features are likely to be closer together than other types of pairs and ii) mutually closest pairs are likely to be formed by pairs from the same class (homogeneous pairs). 
We demonstrate that they provide cues to distinguish normal samples from anomalies when the variance of normal data is smaller than that of the anomalies.

We summarize the contributions of our work as follows:
\begin{itemize}[noitemsep,topsep=0pt,left=0pt]
\item 
A previously-untouched statistical analysis of pairwise distance of features which  provides a cue to distinguishing confident normal samples from unlabeled data,

\item 
a new pseudo-labeling approach based on iteratively re-constructed memory bank~(IRMB) designed to utilize above statistics of pairwise distances,

\item
a novel \emph{mutual smoothness} loss which reduces the ill-posedness by aligning anomaly scores of mutually closest feature pairs under the validated assumption that they largely belong to the same class, and

\item 
a simple yet effective iterative learning-based framework for fully unsupervised anomaly detection, achieving state-of-the-art (SOTA) performance in anomaly detection and localization across various contaminated settings on public industrial datasets (MVTec AD and VisA).

\end{itemize}

\section{Related work}
\label{sec:related_work}
We briefly review studies in anomaly detection that are mostly relevant to our work.
\paragraph{One-class anomaly detection}
In the one-class classification setting, we assume the training data contains correctly labeled normal samples and no anomalies.
While a plethora of different methods exist, these can be largely categorized into  i) reconstruction-based methods~\cite{zavrtanik2021draem, zhang2023destseg, zavrtanik2022dsr, shin2023anomaly} which learn to reconstruct a normal image from an anomaly sample and detect the anomaly region via difference of images, ii) embedding-based methods~\cite{roth2022towards, liu2023simplenet, rudolph2022fully, bae2023pni, defard2021padim, deng2022anomaly, cohen2020sub, tien2023revisiting, wang2021glancing} which measure similarity against normal features extracted from a pretrained network, and iii) self-supervised methods~\cite{li2021cutpaste, schluter2022natural, tien2023revisiting} based on generating pseudo-anomalies.

We summarize previous works partially incorporated into our approach, but used differently as detailed in Sec.~\ref{sec:proposed_method}.
The first relevant work is PatchCore~\cite{roth2022towards}, which utilizes a pretrained feature extractor to obtain normal patch features from training data, subsamples them and stores them in a static memory bank.
During inference, query image features are extracted and compared to the memory bank via nearest-neighbor search for anomaly detection.

The second relevant work is SimpleNet~\cite{liu2023simplenet}, which uses self-supervised learning by adding Gaussian noise to normal features to create pseudo anomalies for training alongside normal samples.
These pseudo anomalies are generated in feature space and used to train a simple discriminator network to detect anomalies.
Nevertheless, these two approaches are designed to work with clean normal samples, which is not robust to noisy training data.

\vspace{-3mm}
\paragraph{Fully unsupervised anomaly detection} 
More recently, several studies explored fully-unsupervised learning for industrial anomaly detection whereby the training data is unlabeled and may comprise anomalies.
Most research~\cite{yoon2022selfsupervise, xi2022softpatch, mcintosh2023inter} attempts to eliminate pseudo-anomalies from the training data and re-deploy one-class anomaly detection~\cite{roth2022towards, li2021cutpaste} on the filtered training set.
Xi \etal~\cite{xi2022softpatch} proposed to filter the training data via thresholding based on the value of local outlier factor (LOF)~\cite{breunig2000lof} and re-deploy PatchCore~\cite{roth2022towards} on the filtered dataset using reweighted anomaly scores.
McIntosh and Albu~\cite{mcintosh2023inter} extracted high-confidence normal patches based on the assumption that normal patch features exhibit high span (large in numbers) and low spread (small diversity), and used them to detect anomaly patches.
While both approaches achieve fully unsupervised training, they solely rely on a pretrained feature extractor for constructing a memory bank, so any incorrectly classified anomaly sample can be stuck inside the memory bank and consistently degrade the detection accuracy. 

\begin{figure*}[t]
  \centering
  \subfloat[Synthetic (isotropic Gaussians)]{
    \includegraphics[width=0.26\textwidth]{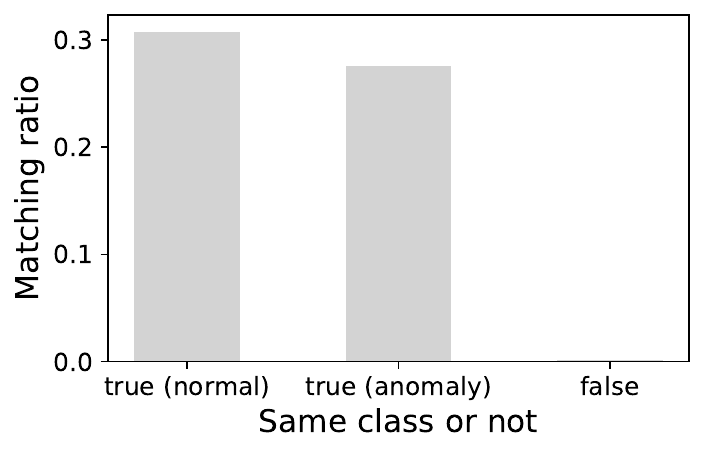}
    \label{fig:matching_ratio_synthetic}
  }
  \hfil
  \subfloat[Real image features (MVTec)]{
    \includegraphics[width=0.26\textwidth]{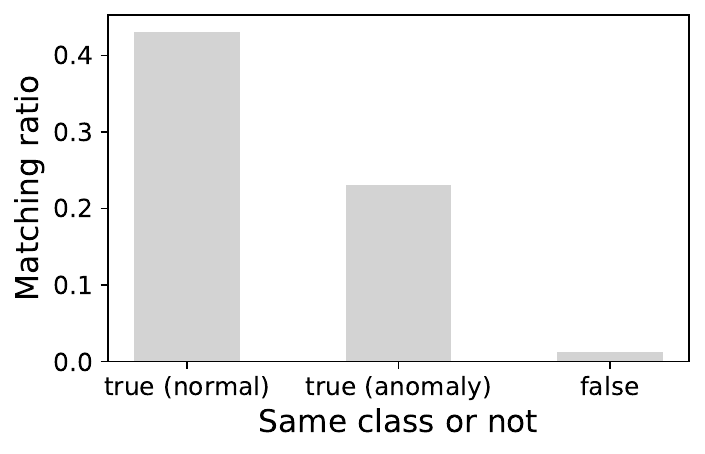}
    \label{fig:matching_ratio_real}
  }
  \hfil
  \subfloat[Real patch features (MVTec)]{
    \includegraphics[width=0.26\textwidth]{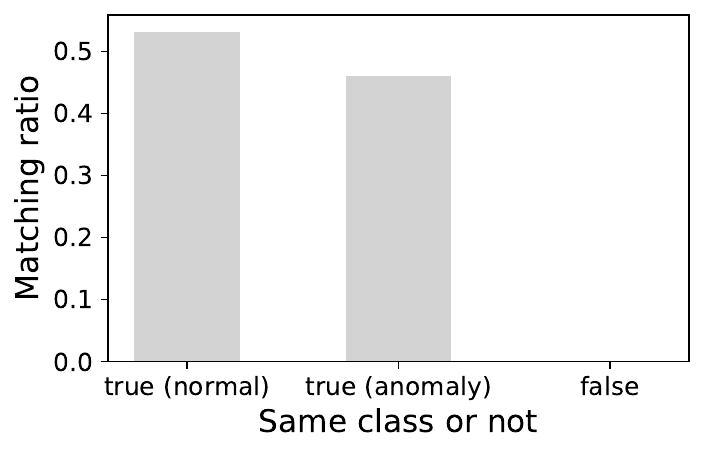}
    \label{fig:matching_ratio_real_patch}
  }
  \vspace{-3mm}
  \caption{
    Visualization of the matching ratios for different types of feature pairs. 
    The empirical experimental settings are as in Sec.~1 in ~\cite{supmat} for both synthetic and real.
    For ``true (normal)'', the matching ratio is the number of samples mutually closest to normal-normal divided by the number of normal samples, and for ``true (anomaly)'', the matching ratio is the number of samples mutually closest to anomaly-anomaly divided by the total number of anomaly samples. Additionally, ``false'' represents the number of samples mutually closest to normal-anomaly (or anomaly-normal) divided by the total number of samples (including both normal and anomaly).
  }
  \label{fig:matching_pairwise_dist}
  \vspace{-2mm}
\end{figure*}

\section{Motivations}
\label{sec:motivation}
We illustrate two observations motivating our strategy proposed in Sec.~\ref{sec:proposed_method}.
In Sec.~\ref{sec:statistical_analysis_pairwise_dist}, we show analytically that the pair of features that are relatively close is most likely to arise from the pair of normal samples.
In Sec.~\ref{sec:smoothness_motivation}, we empirically demonstrate that the mutually closest feature pairs are highly likely to be derived from the same class.
Interestingly, these results only rely on the assumption of smaller variance for the normal data compared to the anomalies, and they do not require the means of two distributions to be different or need anomalies to be scarce.

\begin{figure*}
\centering
\includegraphics[width=0.93\textwidth]{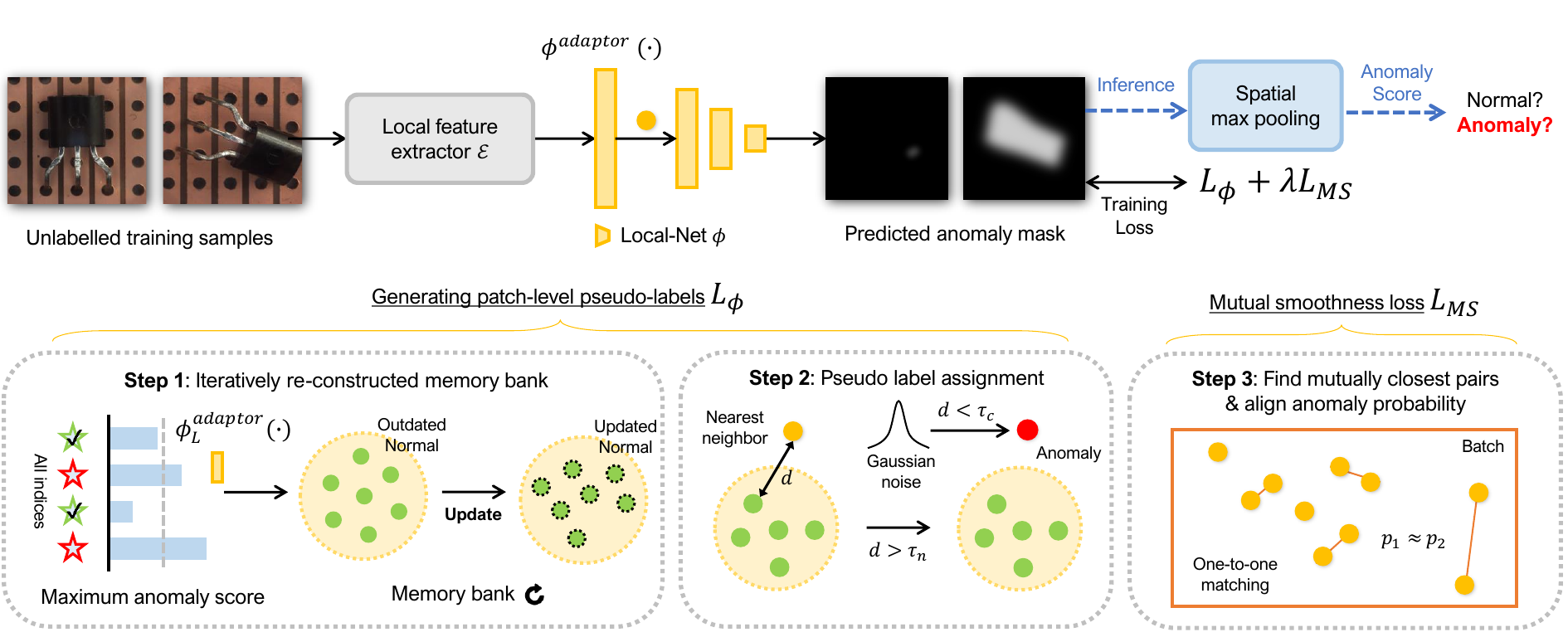} 
\vspace{-3mm}
\caption{
    Our framework overview for fully unsupervised anomaly detection.
    While the framework itself is simple, its constituent components such as patch-level pseudo-label generator and mutual smoothness loss are designed to effectively utilize the observations in Sec.~\ref{sec:motivation}.
}
\vspace{-2mm}
\figlabel{architecture}
\end{figure*}

\subsection{Statistical analysis of pairwise distances between features}
\label{sec:statistical_analysis_pairwise_dist}

For the purpose of intuitive illustration, we present our analysis to an ideal case whereby the normal features and anomaly features follow distinct isotropic Gaussian distributions.
Let $\v x_N \sim \mathcal{N}(\boldsymbol\mu_N, \m \Sigma_N)$ be the distribution of the normal samples with $\m \Sigma_N = \sigma^2_N \m I$ and $\v x_A \sim \mathcal{N}(\boldsymbol\mu_A, \m \Sigma_A)$ the distribution of the anomaly samples with $\m \Sigma_A = \sigma^2_A \m I$.
Provided that the anomaly samples are more spread out than the normal samples, we acknowledge $\sigma_N < \sigma_A$.

If $\v x_1$ and $\v x_2$ are samples each drawn from one of the two distributions, 
then it naturally follows that $\v u := \v x_1 - \v x_2 \sim \mathcal{N}(\boldsymbol\mu_{c_1} - \boldsymbol\mu_{c_2}, (\sigma_{c_1}^2 + \sigma_{c_2}^2) \m I)$, where $c_1$ is the class ($N$ or $A$) of sample 1 and $c_2$ is the class of sample 2 respectively.
Then, the probability of the distance between $\v x_1\in\real^D$ and $\v x_2\in\real^D$ being less than the threshold $\tau\in\real^+$ can be represented as:
\begin{align}
P(\|\v x_1 - \v x_2\|_2 < \tau ~|~ c_1, c_2) 
&= P(\|\v u\|_2^2 < \tau^2 ~|~ c_1, c_2).
\eqlabel{pairwise_dist_1}
\end{align}
If $\v x_1$ and $\v x_2$ are from the same distribution, then $\v u \sim \mathcal{N}(\v 0, 2 \sigma_{c_1}^2 \m I)$, and thus Eq.~\eqref{pairwise_dist_1} becomes 
\begin{align}
P\left(\sum_{i=1}^D u_i^2 < \tau^2 ~|~ c_1 = c_2\right) &= 
P\left(\sum_{i=1}^D z_i^2  < \frac{\tau^2}{2 \sigma_{c_1}^2} \right) \nonumber\\
&= F_{\chi} \left( \frac{\tau^2}{2 \sigma_{c_1}^2} \right),
\eqlabel{pairwise_dist_same_class}
\end{align}
where $\v z \sim \mathcal{N}(\v 0, \m I)$ and $F_\chi(\cdot)$ is the cumulative distribution function of the chi-squared distribution $\chi_D^2$ with $D$ degrees of freedom.
If $\v x_1$ and $\v x_2$ are drawn from different distributions, 
\begin{align}
P\left(\sum_{i=1}^D u_i^2 < \tau^2 ~|~ c_1 \neq c_2\right) 
&= P\left(\sum_{i=1}^D v_i^2 < \frac{\tau^2}{\sigma_N^2 + \sigma_A^2} \right) \nonumber\\
&= F_{\tilde{\chi}} \left( \frac{\tau^2}{\sigma_N^2 + \sigma_A^2} \right),
\eqlabel{pairwise_dist_diff_class}
\end{align}
where $\v v \sim \mathcal{N}((\boldsymbol\mu_{c_1} - \boldsymbol \mu_{c_2}) / \sqrt{\sigma_N^2 + \sigma_A^2}, \m I)$ and $F_{\tilde\chi}(\cdot)$ is the cumulative distribution function of the non-central chi-squared distribution with the non-centrality parameter 
$\lambda = |\boldsymbol\mu_{c_1} - \boldsymbol \mu_{c_2}|^2 / (\sigma_N^2 + \sigma_A^2)$.

We now analyze the probability in Eq.~\eqref{pairwise_dist_1} for different types of features pairs, namely the normal-normal pair, anomaly-anomaly pair and normal-anomaly pair, with $\tau \ll 1$ to simulate close pairs.
Since Eq.~\eqref{pairwise_dist_1} tends to 0 as $\tau\rightarrow0$ for any type of feature pairs, we instead resort to evaluating the ratio of probabilities between different types of pairs to approximate the comparative sizes.
Comparing the probabilities between the normal-normal pair and the anomaly-anomaly pair using Eq.~\eqref{pairwise_dist_same_class} yields
\begin{align}
\frac{P(\|\v x_1 - \v x_2\|_2 < \tau ~|~ c_1=c_2=N)}{P(\|\v x_1 - \v x_2\|_2 < \tau ~|~ c_1=c_2=A)}
& = \frac{F_\chi \left( \tau^2 / 2 \sigma_N^2 \right)}{F_\chi \left( \tau^2 / 2 \sigma_A^2 \right)} > 1
\eqlabel{pairwise_dist_nn_vs_aa}
\end{align}
for $\tau \ll 1$ since $\sigma_N^2 < \sigma_A^2$.
This means that a pair of normals is \emph{more likely} to be within the distance of $\tau$ than a pair of anomalies.
We compare the probabilities between the normal-normal pair and the normal-anomaly pair for $\tau \ll 1$, yielding
\begin{align}
\frac{P(\|\v x_1 - \v x_2\|_2 < \tau ~|~ c_1=c_2=N)}{P(\|\v x_1 - \v x_2\|_2 < \tau ~|~ c_1 \neq c_2)}
& = \frac{F_\chi \left( \tau^2 / 2 \sigma_N^2 \right)}{F_{\tilde\chi} \left( \tau^2 / (\sigma_N^2 + \sigma_A^2) \right)}.
\eqlabel{normal-anomaly_pair}
\end{align}
We consider two lower-bound cases to show that Eq.~\eqref{normal-anomaly_pair} is always greater than 1.
First, when the two distributions have the same mean, i.e. $\boldsymbol \mu_N = \boldsymbol \mu_A$, then $F_{\tilde\chi} ( \tau^2 / (\sigma_N^2 + \sigma_A^2))$ approximates to $F_\chi(\tau^2 / (\sigma_N^2 + \sigma_A^2))$ which is less than $F_\chi(\tau^2 / 2 \sigma_N^2)$ so long as $\sigma_N < \sigma_A$, yielding Eq.~\eqref{normal-anomaly_pair} to be greater than 1.
Second, when the normal and anomaly distributions have substantially different means but similar variances, then $F_{\tilde\chi} ( \tau^2 / (\sigma_N^2 + \sigma_A^2)) \approx F_{\tilde\chi} (\tau^2 / 2\sigma_N^2)$ which cannot be larger than $F_\chi (\tau^2 / 2\sigma_N^2)$ for $\lambda > 0$.
In practice, the normal and anomaly data usually have different means and variances to safely go over 1.
This implies a pair of normal features is \emph{more likely} to be within the distance of $\tau$ than a heterogeneous pair of anomaly and normal features.

\subsection{Empirical analysis on mutually closest features}
\label{sec:smoothness_motivation}
This section is motivated by the seminal work of Zhou \etal~\cite{zhou2003learning}, which leverages the assumption that nearby points are likely to belong to the same class for semi-supervised learning.
Similarly, we turn our attention to analyzing the type of feature pairs formed between mutually closest pairs and aim to check if this label-consistency assumption can be applied to our problem.
Since the statistical analysis of mutually exclusive pairs is complex, we directly resort to empirical analysis to classify the type (heterogeneous or homogeneous) of these pairs.
As in the empirical validation part of Sec.~1 in ~\cite{supmat}, we perform this analysis on the synthetic data comprising the same isotropic Gaussian distributions and real data from MVTec AD~\cite{bergmann19mvtec}.

In the synthetic experiment, we identified the closest sample for each data point and counted the instances where they formed mutually closest pairs. The matching ratio was then calculated as twice the number of unique mutually closest pairs divided by the total number of participating samples. For example, the matching ratio for normal-normal pairs is determined by doubling the number of mutually closest normal-normal pairs and dividing it by the total number of normal samples. As shown in Fig.~\figref{matching_ratio_synthetic}, nearly all mutually closest pairs were homogeneous.

In the real-world experiment, the same process for calculating the matching ratio was applied separately using image-level features and patch-level features for each sequence in the MVTec AD dataset~\cite{bergmann19mvtec}. The results for both feature levels were averaged to produce Figs.~\figref{matching_ratio_real} and~\figref{matching_ratio_real_patch}, which indicate a low proportion of heterogeneous mutually closest pairs. These findings underscore the importance of enforcing class consistency between closest pairs.

\section{Proposed method}
\label{sec:proposed_method}

We describe a learning framework called \textit{FUN-AD} for fully unsupervised industrial anomaly detection, which consists of an anomaly pseudo-labeling method motivated by Sec.~\ref{sec:statistical_analysis_pairwise_dist} and a loss function inspired by Sec.~\ref{sec:smoothness_motivation}.
\vspace{-4mm}

\paragraph{Preliminaries}
We define the training set as $\mathcal{X}:=\{ I_i \}_{i=1}^N$, where $I_i\in\real^{H \times W \times 3}$ is the $i$-th image, $N$ is the number of training samples, and $H$ and $W$ are the image height and width, respectively. We define the $j$-th patch of $I_i$ as $X_{ij}\in\mathbb{R}^{K\times K \times 3}$, where $K$ is the patch size. 
FUN-AD comprises two sub-networks: a feature extractor $\mathcal{E}$ and the Local-Net model $\phi$ for detecting anomalies. $I_i$ is passed through $\mathcal{E}$ to extract the patch-level features $\{{\v f_{ij}}\}_{j=1}^P$.

\subsection{Generating patch-level pseudo-labels from pairwise-distance statistics}
\label{sec:nn_search}

From Sec.~\ref{sec:statistical_analysis_pairwise_dist}, we note that feature pairs with smaller pairwise distances are more likely to be homogeneous normal pairs, provided that the anomaly features are more spread out than normal features. 
This observation motivates us to utilize the statistics for pseudo-labeling.

We update the patch feature vector of images classified as normal by $\phi$ inside the memory bank at each iteration.
This implies that even with a randomly constructed (noisy) memory bank containing as many anomalies as normal samples, analyzing the statistics of pairwise distances will allow us to distinguish some confident normal and anomalous samples from the unlabeled training set, providing sufficient supervision to initiate the learning process.

This approach demonstrates that even in the early stages of training, when the memory bank is nearly random, it predominantly consists of normal samples.
Since some of the normals in the initial memory bank will pull other normals into memory and push out anomalies, only normals will remain in the memory bank after an iteration. 
The normal-only memory bank will no longer be noisy and will therefore be better able to distinguish normal from abnormal.
Hence, we propose to gradually refine our memory bank features through iteratively re-constructed memory banks and assign pseudo-labels based on pairwise distances. 

\vspace{-3mm}
\paragraph{Iteratively re-constructed memory bank (IRMB)} 
In each iteration, we construct a memory bank comprising features likely derived from normal images. 
To achieve this, we first estimate the global anomaly score of each image by max-pooling the patch-level anomaly scores of the constituent patches, i.e., $\max_j\phi(\v f_{ij})$. 
Then, we apply min-max normalization to these scores across all training images and use a threshold $\tau_b$ to identify a set $\mathcal{P}$ comprising features that are more likely to be normal.
Anomaly scores from the local network are normalized, but since they are mostly distributed near 0.5 at the beginning, we perform min-max normalization to distinguish between confident normal and confident abnormal.
In terms of equation,

\begin{align}
    \mathcal{P} = \left\{i \;\middle|\; \frac{\max_j\phi(\v f_{ij})-\min_i \max_j\phi(\v f_{ij})}
    {\max_i \max_j\phi(\v f_{ij})-\min_i \max_j\phi(\v f_{ij})} < \tau_b \right\}.
\label{eq:index_collection}
\end{align}
Additionally, we sample a random subset $\mathcal{P'} \subset \mathcal{P}$ to reduce computational time.
Finally, we construct a memory bank $\mathcal M = \{ \phi^{adaptor}(\v f_{ij}) ~~|~~ i\in\mathcal P', j=1,\cdots,P \}$ by storing patch-level features from $\mathcal{P'}$ that have additionally passed through the learnable feature adaptor of the Local-Net ($\phi^{adaptor}$).
This allows features from pretrained $\mathcal{E}$ adapt to our anomaly detection task.
This feature adaptation along with iteratively reconstructed memory bank allows gradually sharpening of the learning signal.

\begin{figure}[t]
\begin{algorithm}[H]
\small
\caption{Training procedure for \textit{FUN-AD}}\label{training_algorithm}
\begin{algorithmic}[1]
\Require Unlabeled training set $\mathcal{X}:=\{ I_i \}_{i=1}^N$.
\Ensure Weights of the Local-Net $\phi$.

\For{$i=1,\dots,N$}{}
    \State Extract local patch features $\{\v f_{ij}\}_{j=1}^P$ from image $I_i$.
\EndFor
\For{$m=1,\dots,\mathtt{num\_epochs}$}{}
    \State \textbf{\textit{Local-Net Training.}}
    \State Divide $\{\v f_{ij}\}_{j=1}^P$ for $i=1,\cdots,N$ into mini-batches $\{\mathcal{B}_n\}$.
    \For{$n=1,\dots,\mathtt{num\_iters}$}{}
        \State Construct $\mathcal{M}$ using $\phi$, Eq.~(6).
        \State Compute $s_{ij}~\forall~\v f_{ij}\in\mathcal{B}_n$ using Eq.~(7) and Eq.~(8).
        \State Compute hard labels $y_{ij} \leftarrow H(s_{ij} - \tau_n) ~\forall~\v f_{ij}\in\mathcal{B}_n$.        
        \State Find mutually closest feature pairs ($\v {f}_{ij}, \v {f}_{kl}$) within the mini-batch based on Eq.~(10).
        \State Compute mutual smoothness loss using Eq.~(9)
        \State Augment feature $\v f_{ij} \leftarrow \v f_{ij}+\epsilon~\forall~\v f_{ij} \in \mathcal{F}$ (See Sec.4.3)
        \State Perform 1 iteration of nonlinear optimization to minimize $L_\phi + \lambda L_{MS}$.
    \EndFor
\EndFor
\end{algorithmic}
\label{alg:training}
\end{algorithm}
\vspace{-9mm}
\end{figure}
\begin{figure*}[!t]
  \centering
  \subfloat[Initial anomaly scores]{
    \includegraphics[width=0.26\textwidth]{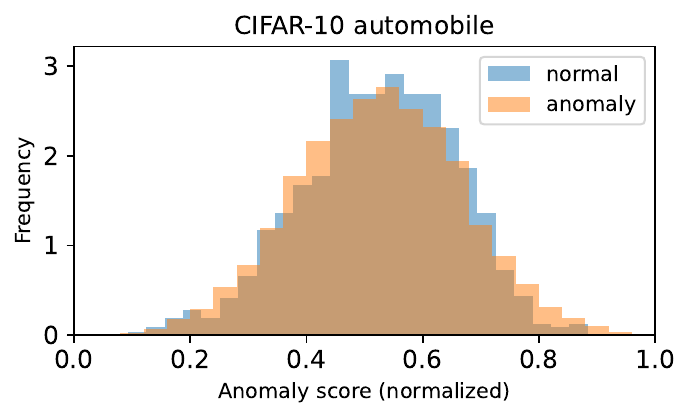}
    \label{fig:initialization_hist}
  }
  \hfil
  \subfloat[Final anomaly scores]{
    \includegraphics[width=0.26\textwidth]{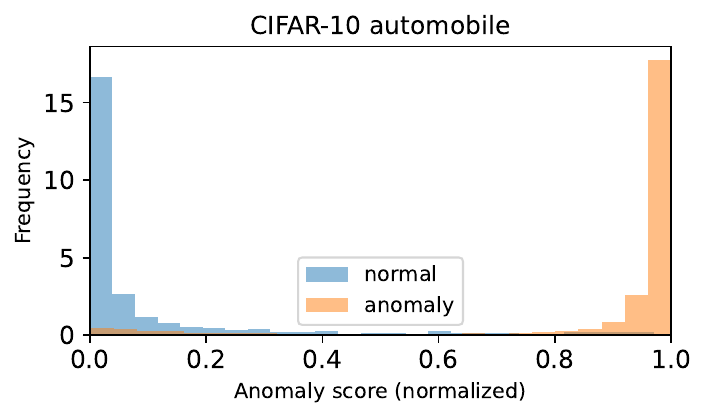}
    \label{fig:training_hist}
  }
  \hfil
  \subfloat[AUROC per iteration]{
    \includegraphics[width=0.26\textwidth]{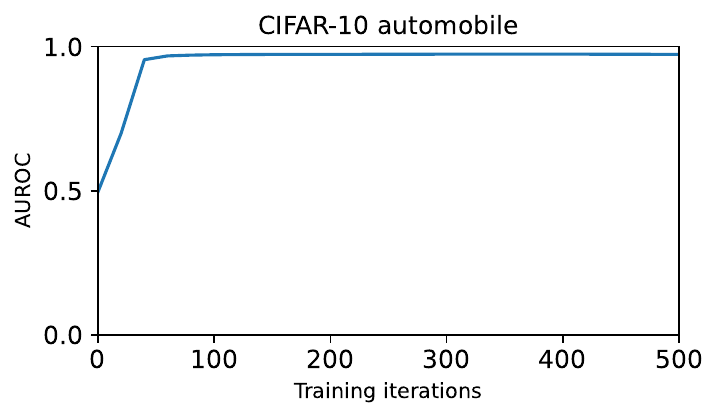}
    \label{fig:auroc_graph}
  }
  \caption{
    Illustration of FUN-AD before and after training on a semantic anomaly detection example.
    The histograms of anomaly scores from the Local-Net before and after training are shown in (a) and (b) respectively.
    (c) shows evolution of the image-wise detection accuracy over the iterations.
  }
  \label{fig:semantic_anomaly_results}
  \vspace{-5mm}
\end{figure*}

\vspace{-3mm}
\paragraph{Pseudo-label assignment}
In each iteration, we utilize the pairwise distance statistics linked to IRMB for assigning patch-level pseudo labels. 
We conduct a nearest-neighbor search for each adapted patch feature against the internal features of the memory bank $\mathcal{M}$, excluding the feature itself.
This exclusion is necessary because the initial Local-Net is random, and not all feature vectors in the memory bank can reliably be considered normal. 
If a query feature inside the memory bank requires pseudo-labeling, it may continue to be labeled as normal, resulting in persistent incorrect pseudo-labels in the absence of exclusion.

Initially, we define the patch features within the minibatch of the $n$-th iteration as $\mathcal B_{n} =\{ \{ \v f_{ij}\}_{j=1}^P\}_{i=1}^B$, where $B$ is the batch size.
The nearest-neighbor distance $d_{ij}$ to the features in $\mathcal{M}$ is defined as:
\begin{align}
\label{eq:distance} 
d_{ij} &= \min \Big( \{\norm{\phi^{adaptor}(\mathbf{f}_{ij}) - \v m}_2 : \v m \in \mathcal{M} \nonumber\\
& \qquad\qquad \land \norm{\phi^{adaptor}(\mathbf{f}_{ij}) - \v m}_2 > 0\} \Big),
\end{align}
where $\land$ is an \textit{and} operator to remove the case where $\v m$ is derived from $\v f_{ij}$.
This distance is min-max normalized to yield the anomaly score for pseudo labeling ($s_{ij}$) as:
\begin{align}
\label{eq:normalization}
    s_{ij} = \frac{d_{ij}-\min_{i,j} d_{ij}}{\max_{i,j} d_{ij} - \min_{i,j} d_{ij}},
\end{align}
which is thresholded to assign the patch-level pseudo label as $y_{ij} = H(s_{ij} - \tau_n)$, where $H(x)$ is a unit step function and $\tau_n$ is the threshold below which is classified as normal. 
Since our goal is to distinguish between outliers and normals, we use a threshold to divide the regions taking advantage of the large spikes in maximum values caused by outliers.
This assignment strategy, based on pairwise-distance statistics, is robust to the initially random memory bank and can still provide correct learning signals from the confident normal and anomaly samples. 
The remaining issues with false positives and false negatives are addressed in Sec.~\ref{sec:training}.

\subsection{Mutual-smoothness loss}
\label{sec:local_smoothness}
Following the pseudo-label assignment process, many incorrect pseudo-labels may result from assigning hard labels based on a threshold, potentially leading to inaccurate learning.
Therefore, building on the observation from Sec.~\ref{sec:smoothness_motivation} that the mutually closest pairs of features are likely to share the same class, we propose a new \emph{mutual smoothness loss} to align the patch-level anomaly scores of the features forming a mutually-closest pair.
For this purpose, we employ the $L_1$ loss function known for its robustness against noisy labels as outlined in \cite{ghosh2017robust}, yielding
\begin{align}
\mathcal{L}_{MS} = \frac{1}{|\Omega|} \sum_{(ij,kl)\in\Omega} |\phi(\v {f}_{ij}) - \phi(\v {f}_{kl})|,
\label{eq:121_loss}
\end{align}
where $\Omega$ is defined as the unique set of mutually closest pairs such that, for all $(ij,kl) \in \Omega$,
\begin{align}
& \min_{i \neq k, j \neq l} \|\phi^{\text{adaptor}}(\v {f}_{ij}) - \phi^{\text{adaptor}}(\v {f}_{kl})\|_2  \nonumber \\
&\qquad\qquad= \min_{k \neq i, l \neq j} \|\phi^{\text{adaptor}}(\v {f}_{ij}) - \phi^{\text{adaptor}}(\v {f}_{kl})\|_2.
\label{eq:closest_pair}
\end{align}
In ambiguous situations where the pseudo-label scores of samples that are mutual nearest-neighbors are close to the threshold, each sample may be assigned a different label.
However, since mutually-nearest-neighbor pairs are likely to be in the same class, the anomaly scores between the two samples are made similar to prevent incorrect prediction.
The positive effect of this loss is demonstrated in Table~\ref{tab:msl_syn_ablation}.

\subsection{Training procedure}
\label{sec:training}
Subsequently, the Local-Net is trained to minimize the total loss function $\mathcal{L} = {L}_{\phi}+\lambda{L}_{MS}$, which is a weighted sum of the balanced cross-entropy loss ${L}_{\phi}$ and the mutual smoothness loss $L_{MS}$.
${L}_{\phi}$ is defined as
\begin{align}
\mathcal{L}_{\phi} &= \frac{1}{|\mathcal{B}_{A,n}|} \sum_{(i,j) \in \mathcal{B}_{A,n}} y_{ij} \ln \phi(\v f_{ij}) \nonumber\\
&\quad + \frac{1}{|\mathcal{B}_{N,n}|} \sum_{(i,j) \in \mathcal{B}_{N,n}} (1-y_{ij}) \ln (1 - \phi(\v f_{ij}))
\label{localnet_loss}
\end{align}
based on the pseudo-labels assigned from Sec.~\ref{sec:nn_search}, and the mutual smoothness loss expressed in Sec.~\ref{sec:local_smoothness}. $\mathcal{B}_{A,n}$ and $\mathcal{B}_{N,n}$ are the set of samples pseudo-labeled as anomaly and normal respectively in iteration $n$.
\vspace{-3mm}
\paragraph{Augmentation of ambiguous features via Gaussian perturbation}
\label{sec:false_positive}
The hard pseudo-labels from Sec.~\ref{sec:nn_search} inevitably yield false positives and false negatives. While it is difficult to avoid them completely, we introduce simple feature augmentation to reduce their negative impact. 
This is achieved by adding Gaussian noise to the set of ambiguous features, which are classified as anomalies, but do not acquire scores above the confident-anomaly threshold of $\tau_c$.
This approach mitigates the problem by introducing noise as a perturbation, which prevents the model from incorrectly classifying normal features as anomalies.
Our method is partly motivated by~\cite{qiu2022latent, Du2020Robust, liu2023simplenet}, which demonstrate that additive Gaussian noise applied to the normal features can generate useful pseudo-anomalies.
Above can be expressed as $\v f_{ij} \leftarrow \v f_{ij} + \v{\epsilon}~\forall~\v f_{ij} \in \mathcal F$, 
where $\mathcal F = \{ \v f_{ij} ~|~ \tau_n < s_{ij} < \tau_c \}$ is the set of ambiguous anomalies and $\v{\epsilon}\sim\mathcal N(\v{0}, \Lambda)$ is the Gaussian perturbation with $\Lambda$ being a diagonal covariance matrix with the elements computed by estimating the element-wise variance of the patch-level features in the mini-batch, i.e. $\v f_{ij} \in \mathcal{B}_{n}$. Finally, the Local-Net is updated by incorporating all of the aforementioned steps. See Algorithm 1 for detailed model training steps.

\begin{table*}[t]
\centering
\fontsize{7}{8}\selectfont
\renewcommand{\arraystretch}{1.2}
\begin{tabular}{cc|rrrrr|rrr}
\Xhline{2\arrayrulewidth}
\multicolumn{2}{c|}{Type} & \multicolumn{5}{c|}{One-class classification} & \multicolumn{3}{c}{Fully unsupervised} \\ 
Dataset & Method & CS-Flow~\cite{rudolph2022fully} & PaDiM~\cite{defard2021padim} & PatchCore~\cite{roth2022towards} & SimpleNet~\cite{liu2023simplenet} & 
RealNet~\cite{zhang2024realnet} &
SoftPatch~\cite{xi2022softpatch} & InReaCh~\cite{mcintosh2023inter} & FUN-AD (\textit{Ours}) \\

\hline
\multirow{2}{*}{MVTec AD} & \textit{No overlap} & 93.8 / - & 94.5 / 96.6 & 98.1 / 94.7 & 97.7 / 95.3 & \textbf{99.2} / 96.9 &  \underline{98.3} / \underline{97.3} & 92.2 / 96.9 & \textbf{99.2} / \textbf{98.6} \\
\cline{2-10}
 & \textit{Overlap} & 65.5 / - & 70.1 / 91.4 & 81.6 / 70.8 & 59.3 / 52.9 & 96.8 / 96.0 & \underline{98.3} / 95.1 & 92.4 / \underline{97.2} & \textbf{98.8} / \textbf{98.6} \\
\hline
\multirow{2}{*}{VisA} & \textit{No overlap} & 82.3 / - & 85.8 / 98.3 & 89.8 / 95.7 & 90.4 / 96.7 & \underline{93.5} / \underline{98.7} & 89.6 / 98.5 & 83.8 / 97.6 & \textbf{95.0} / \textbf{99.2} \\
\cline{2-10}
 & \textit{Overlap} & 64.3 / - & 70.1 / 92.3 & 82.3 / 81.6 & 50.4 / 57.9 & \underline{92.4} / \underline{97.7} & 89.5 / 96.3 & 78.5 / 93.6 & \textbf{95.1} / \textbf{99.2} \\
\Xhline{2\arrayrulewidth}
\end{tabular}
\vspace{-2mm}
\caption{
    Comparison of quantitative experimental results on the MVTec AD (contaminated) and VisA datasets.
    The table displays image-wise AUROC (\%) / pixel-wise AUROC (\%), representing anomaly detection and localization performance, respectively. The best results are in bold and the runner-ups are underlined.
}
\vspace{-3mm}
\label{tab:iad_main}
\end{table*}

\section{Experimental results and discussions}
\label{sec:experimental_results}

We compared our method against several baselines using industrial anomaly detection benchmark datasets.
We also evaluated the performance when varying the percentage of anomalies in the training dataset, the presence of each module, and the hyperparameters through ablation study.

\subsection{Toy example of semantic anomaly detection}
\label{sec:toy_example}
We used CIFAR-10~\cite{krizhevsky2009learning} to conduct a toy experiment.
The normal class is ``automobile'', and the outliers consist of the remaining classes in CIFAR-10. 
Fig.~\ref{fig:initialization_hist} shows the semantic anomaly scores when Local-Net is randomly initialized, and Fig.~\ref{fig:training_hist} shows the semantic anomaly scores after Local-Net has been trained by FUN-AD's training process. 
Fig.~\ref{fig:auroc_graph} illustrates that the AUROC metric shows a clear separation between normal and anomaly, with most of the normal samples remaining in the memory bank as training progresses.
For more details, please refer to~\cite{supmat}.

\subsection{Experiments}
\label{sec:experiments}

\paragraph{Datasets} We primarily utilized two widely recognized public benchmarks, MVTec AD~\cite{bergmann19mvtec} and VisA~\cite{zou2022spot}. 
MVTec AD comprises 15 categories (10 objects, 5 textures), and VisA includes 12 object categories. 
For MVTec AD, we modified the one-class classification setup by randomly incorporating some of the test set anomalies into the training set at a 1:10 ratio, creating noisy training data contaminated with anomalies.
All training samples were stripped of their labels to construct a fully unsupervised setting.
In the \textit{\textbf{No overlap}} scenario, these relocated anomalies were excluded from evaluation across different anomaly-to-normal ratios. 
In the \textit{\textbf{Overlap}} scenario, the anomalies moved from the test set to the training set were also used for inference. 
We also considered \textit{\textbf{Overlap}} scenario, as existing fully unsupervised anomaly detection baselines~\cite{xi2022softpatch, mcintosh2023inter} have been evaluated.

\vspace{-3mm}
\paragraph{Main results}
In Table~\ref{tab:iad_main}, \textit{FUN-AD} outperforms previous SOTA methods in both anomaly detection and localization on the contaminated MVTec AD in the \textbf{\textit{No overlap}} setting.
In the \textbf{\textit{Overlap}} setting, our model performs almost as well, unlike the degradation observed in one-class classification models, including fully-unsupervised methods.

\begin{figure}[!t]
    \centering
    \includegraphics[width=0.9\columnwidth]{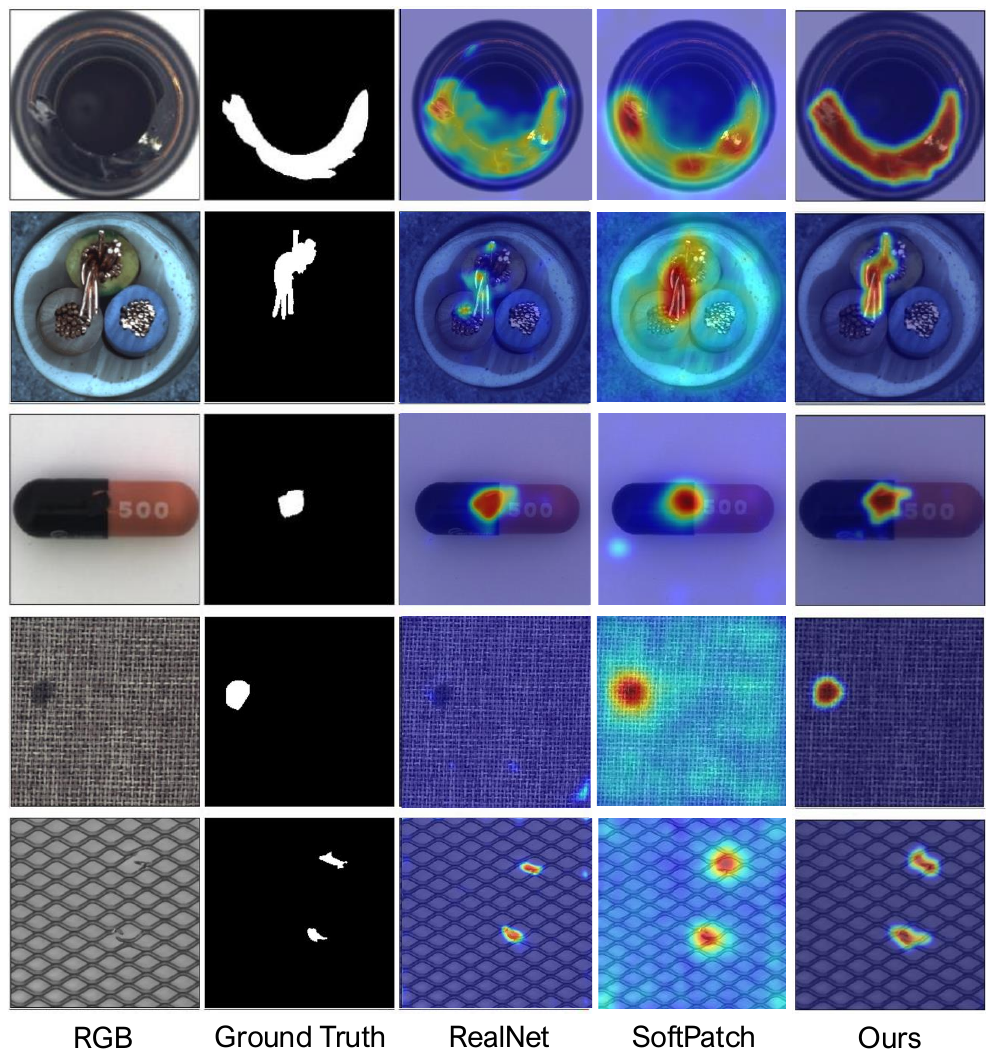}
    \label{fig:qualitative_comparison_mvtec}
    \vspace{-3mm}
    \caption{
         Visualization of anomaly detection results achieved by SOTA models on the MVTec AD dataset. Within each group, from left to right, are the anomaly image, ground-truth, and predicted anomaly scores of each model.
    }
\vspace{-5mm}
\label{fig:qualitative_comparison}
\end{figure}

Similarly, on VisA, our model exhibits significantly improved results compared to existing models.
In the \textbf{\textit{Overlap}} setting of VisA, where other one-class classification models show significant performance degradation, our model maintains consistent accuracy.
In particular, we show that our method achieves robust performance on the VisA dataset, which contains multiple objects and lacks camera alignment. 
Fig.~\figref{qualitative_comparison} shows our method produces sharper boundaries than models that resemble one-class classification.

\subsection{Ablation study}
\label{sec:ablation_study}
In ablation study, we considered the possibility that the training dataset may not be contaminated in real-world scenarios.
We emphasize that the synthetic anomalies were generated using a noisy (anomaly-present) dataset, which differs from the approach in \cite{zhang2024realnet} that requires clean samples, potentially deteriorating the quality of generated images.
We conducted all ablation studies, except those related to semantic anomaly detection, using the training dataset with synthetic anomalies added. 
Additionally, we used MVTec AD (with 10\% contamination) for all ablations except those related to semantic anomaly detection and contamination rate.
Additional ablation studies can be found in the supplementary document~\cite{supmat}.

\begin{table*}[t]
\renewcommand{\arraystretch}{1.2}
\fontsize{7}{8}\selectfont
    \centering
        \begin{tabular}{c|c|cccccc}
            \hline
            \multirow{2}{*}{Dataset} & \multirow{2}{*}{Method} & \multicolumn{6}{c}{Anomaly-to-normal ratio} \\ \cline{3-8} 
            & & $0\%$ & $1\%$ & $3\%$ & $5\%$ & $10\%$ & $20\%$ \\ \hline
            \multirow{4}{*}{MVTec AD~\cite{bergmann19mvtec}} & InReach~\cite{mcintosh2023inter} & 92.43 / 97.07 & 92.40 / 97.17 & 92.40 / 97.11  & 92.32 / 97.01 & 92.17 / 96.89 & 91.28 / 96.94 \\
            & SoftPatch~\cite{xi2022softpatch} & {\bf98.32} / {\bf98.28} & {\bf98.29} / {\bf98.28} & \underline{98.38} / \underline{98.27} & 98.40 / 98.17 & 98.33 / 97.28 & 97.73 / 96.87 \\
            & FUN-AD & 93.33 / 96.19 & 95.11 / 97.54 & 95.99 / 97.52 & \underline{98.45} / \underline{98.43} & {\bf99.23} / {\bf98.55} & {\bf98.41} / \underline{98.15} \\ 
            & FUN-AD* & \underline{95.63} / \underline{97.51} & \underline{98.11} / \underline{98.19} & {\bf98.51} / {\bf98.37} & {\bf98.85} / {\bf98.49} & \underline{98.95} / \bf{98.55} & \underline{98.36} / {\bf98.35} \\ \hline

            \multirow{4}{*}{VisA~\cite{zou2022spot}} & InReach~\cite{mcintosh2023inter} & 83.84 / 97.61 & 83.96 / 97.65 & 84.24 / 97.66 & 83.66 / 97.67 & 83.72 / 97.58 & 76.15 / 97.22 \\
            & SoftPatch~\cite{xi2022softpatch} & \underline{90.23} / {\bf98.59}  & \underline{90.08} / \underline{98.56} & 90.02 / 98.59 & 89.82 / 98.57 &  89.69 / 98.58 & 89.17 / 98.48 \\
            & FUN-AD & 87.89 / 98.22  & 89.54 / 98.50 & \underline{91.63} / \underline{98.70}  & \underline{93.65} / \bf{98.92} & \underline{94.57} / {\bf99.13} & \underline{94.50} / \underline{99.07} \\
            & FUN-AD* & {\bf90.71} / {\underline{98.53}}  & {\bf91.85} / {\bf98.61} & {\bf93.35} / {\bf98.90}  & {\bf94.25} / \underline{98.88} & {\bf94.59} / {\bf99.13} & {\bf94.53} / {\bf99.08} \\ \hline
        \end{tabular}
    \vspace{-2mm}
    \caption{Performance comparison of different fully-unsupervised anomaly detection methods across different anomaly-to-normal ratios on the contaminated MVTec AD and VisA datasets (\textit{no overlap}). * indicates synthetic anomaly data has been utilized for training. The best results are in bold and the runner-ups are underlined.}
    \vspace{-2mm}    
    \label{tab:noise_ratio}
\end{table*}

\vspace{-3mm}
\paragraph{Effects of Gaussian noise and mutual smoothness loss}
\label{sec:gaussian_noise_ablation}
Table~\ref{tab:msl_syn_ablation} presents the results of mutual-smoothness loss, and feature augmentation with Gaussian noise.
The comparison with and without $\epsilon$ in Table~\ref{tab:msl_syn_ablation} demonstrates the effectiveness of the feature augmentation approach introduced in Sec.~\ref{sec:training}. 
FUN-AD exhibits a significant performance drop without additive Gaussian noise, highlighting the importance of reducing ambiguous anomaly features when providing effective pseudo-anomalies.
We demonstrate in Table~\ref{tab:msl_syn_ablation} that mutual smoothness loss performs effectively even in the absence of feature augmentation generated by Gaussian noise, a condition that can lead to many false positives.

\begin{table}[t]
    \centering
    \resizebox{0.7\columnwidth}{!}{
        \begin{tabular}{cc|cc} \toprule
            $\mathcal{L}_{MS}$ & $\epsilon$ & $\text{AUROC}_{\textit{\text{image}}}(\%)$ & $\text{AUROC}_{\textit{\text{pixel}}}(\%)$ \\ \midrule
            & & 94.95 & 97.87 \\
            \checkmark & & 96.15 & 97.87 \\
            \checkmark & \checkmark & {\bf 98.95} & {\bf 98.55} \\
            \bottomrule
        \end{tabular}
    }
    \vspace{-2mm}
    \caption{Ablation study of mutual smoothness loss and gaussian noise. $\epsilon$ indicates the addition of Gaussian noise to ambiguous samples as described in Sec.~\ref{sec:training}.} 
    \vspace{-2mm}
    \label{tab:msl_syn_ablation}   
\end{table}
\begin{table}[t]
\renewcommand{\arraystretch}{1}
\centering
\resizebox{0.7\columnwidth}{!}{
\begin{tabular}{c|ccc}
\toprule
       & \multicolumn{3}{c}{$\tau_n$} \\ \midrule
$\tau_c$ & \multicolumn{1}{c|}{0.4} & \multicolumn{1}{c|}{0.5} & 0.6 \\ \midrule
0.8    & \multicolumn{1}{c|}{98.88 / 98.62} & \multicolumn{1}{c|}{\textbf{99.01} / 98.73} & 98.83 / \textbf{98.75} \\ \midrule
0.9    & \multicolumn{1}{c|}{98.73 / 98.33} & \multicolumn{1}{c|}{98.95 / 98.55} &  98.90 / 98.70  \\ \bottomrule
\end{tabular}
}
\vspace{-2mm}
\caption{
    Ablation study of variations in the anomaly threshold $\tau_n$ and confident anomaly threshold $\tau_c$. 
    The table presents the image-wise AUROC (\%) / pixel-wise AUROC (\%) for assessing anomaly detection and localization performance.
}
\vspace{-3mm}
\label{tab:sensitivity_threshold}
\end{table}

\vspace{-5mm}

\paragraph{Sensitivity to hyperparameters $\tau_n$ and $\tau_c$}
\label{sec:sensitivity_threshold}
We examined the sensitivity of patch-level pseudo-labeling to hyperparameters $\tau_n$ and $\tau_c$. 
As mentioned in Sec.~\ref{sec:false_positive}, if $s_{ij}$ is larger than $\tau_c$, we consider it as a confident anomaly.
On the other hand, if $s_{ij}$ is between $\tau_n$ and $\tau_c$, we consider it as an ambiguous situation, where the distinction between normal and anomalous is unclear, and perturb the features by adding Gaussian noise to treat it as an anomaly.
The results in Table~\ref{tab:sensitivity_threshold} demonstrate robustness within a range of $\pm$0.1 from the baseline values of $\tau_n$ and $\tau_c$. 
This suggests that our method is relatively robust to threshold variations, except when $\tau_c$ is significantly lower than the default value of 0.9, causing a high rate of false positives, or when $\tau_n$ falls substantially below the default value of 0.5, resulting in very few normal samples being labeled as normal.

\vspace{-3mm}
\paragraph{Effect of different contamination rates}
\label{sec:contamination_v2}
Table~\ref{tab:noise_ratio} demonstrates the performance of FUN-AD according to the contamination ratio in the training dataset.
Here, ``FUN-AD'' refers to the results obtained from training with the dataset without synthetic anomalies, while ``FUN-AD*'' refers to the results from training the FUN-AD framework with synthetic anomalies added at a rate of 5\% of the training dataset size.
Synthetic anomalies were created from a noisy (anomaly-present) dataset considering a fully unsupervised setting.
Given that they contain noisy samples, which could potentially degrade the quality of generated data, their utility in the training process may not always be advantageous.

Since FUN-AD relies on pseudo-labeled anomaly samples for detection, it does not perform as well as other baselines when the training dataset contains very few anomalies. 
However, when synthetic anomalies are added, the model effectively learns to distinguish between anomalies and normal samples by pseudo-labeling synthetic anomalies in situations where real anomalies are scarce.

\section{Conclusion}
\label{sec:conclusion}
We have addressed the challenging problem of identifying industrial anomalies without any labeled normal or anomaly data in the fully unsupervised setting whereby the training dataset contains anomalies but the labels are unavailable.
Based on the assumption of wider spread for anomalies, we illustrated analytic and empirical motivations for our methodology, namely that normal-normal feature pairs are more likely to form closer feature pairs, and mutually closest pairs are likely to share the same class labels.
To incorporate these observations, we presented a novel unsupervised anomaly detection framework, which assigns pseudo-labels based on iteratively re-constructed memory bank and pairwise-distance statistics to achieve robustness to initial noisy labels and allow gradual refinement of the learning signals.
We also leveraged the class-consistency of mutually closest features by proposing a new MAE-based mutual smoothness loss for training.
Through extensive experimental evaluations, we demonstrated the competitiveness of our approach across different industrial anomaly benchmarks in presence of contaminated training data.

\vspace{-3mm}
\paragraph{Acknowledgement}
This work was in part supported by the Technology Innovation Program (1415178807, Development of Industrial Intelligent Technology for Manufacturing, Process, and Logistics) funded by the Ministry of Trade, Industry and Energy (Korea), in part by the National Research Foundation of Korea (NRF) grant funded by the Korean government~(No. RS-2023-00302424), and in part by the Institute of Information and communications Technology Planning and Evaluation (IITP) under the artificial intelligence semiconductor support program to nurture the best talents (IITP-2024-RS-2023-00253914) grant funded by the Korean government (MSIT).

\clearpage

{\small
\bibliographystyle{ieee_fullname}
\bibliography{egbib}
}

\appendix
\renewcommand{\thesection}{\arabic{section}}
\renewcommand{\thefigure}{\arabic{figure}}
\renewcommand{\thetable}{\arabic{table}}
\setcounter{section}{0}
\setcounter{figure}{0}
\setcounter{table}{0}

\twocolumn[
    \begin{center}
        \Large\textbf{Supplementary Document for FUN-AD: Fully Unsupervised Learning for Anomaly Detection with Noisy Training Data}
        \vspace{0.5cm}
    \end{center}
]

\begin{figure*}[t]
  \centering
  \subfloat[Synthetic (isotropic Gaussians)]{
    \includegraphics[width=0.26\textwidth]{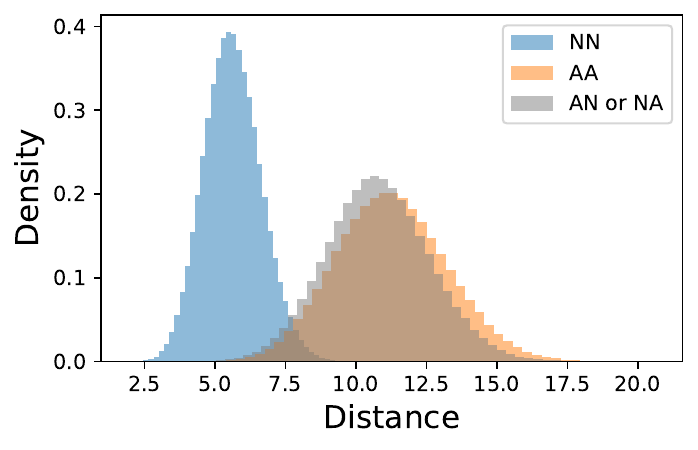}
    \label{fig:synthetic_distance_histogram}
  }
  \hfil
  \subfloat[Real (bottle, image features)]{
    \includegraphics[width=0.26\textwidth]{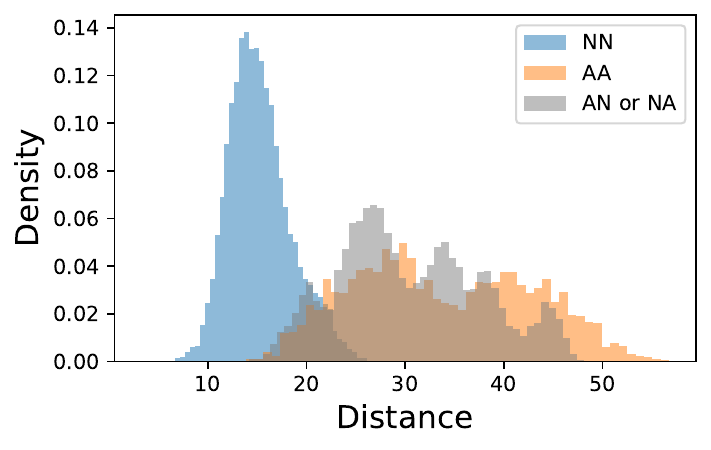}
    \label{fig:real_distance_histogram}
  }
  \hfil
  \subfloat[Real (bottle, patch features)]{
    \includegraphics[width=0.26\textwidth]{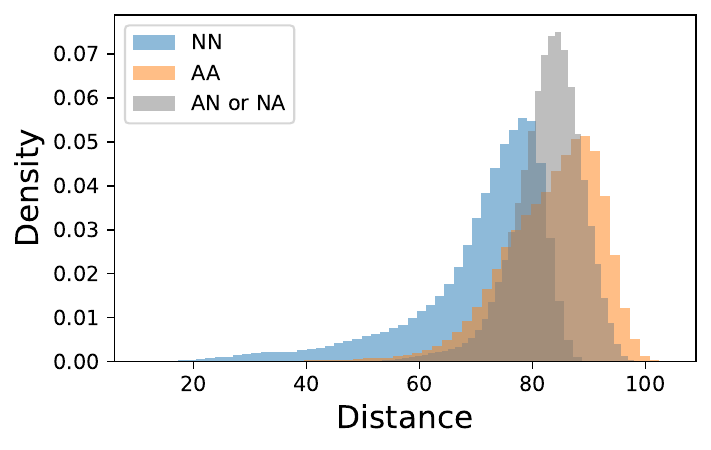}
    \label{fig:real_patch_distance_histogram}
  }
  \vspace{-3mm}
  \caption{
    Histogram of pairwise distances for different types of feature pairs. Abbreviations are as follows: NN for normal-normal pairs, AA for anomaly-anomaly pairs, AN or NA for anomaly-normal or normal-anomaly pairs.
    For the synthetic experiment, the normal samples were drawn from $\mathcal{N}(\v 0, \m I)$ and the anomaly samples from $\mathcal{N}(\v {1.5}, 2 \m I)$.
    For the ``real'' comparison, we used patch features (patch tokens) and image features (class tokens) extracted from the pretrained DINO model~\cite{caron2021emerging} for the bottle set~\cite{bergmann19mvtec}.
  }
  \figlabel{pairwise_dist}
\end{figure*}

\section{Additional statistical analysis of pairwise distances between features}

\paragraph{Empirical validation}
Since our statistical analysis is limited to isotropic Gaussian distributions, it is not directly applicable other distributions or real-world data. 
Therefore, we aim to bridge this theoretical gap with empirical analysis using real-world data. 
We validate these findings on both synthetic data with isotropic Gaussian distributions and real data from the \emph{bottle} set in MVTec AD~\cite{bergmann19mvtec}, utilizing normal images from the training set and anomaly images from the test set, as the training set does not contain any anomalies.

For the synthetic experiment, we sampled 1000 16-dimensional features from the normal distribution $\mathcal{N}(\v 0, \m I)$ and 1000 samples from the anomaly distribution $\mathcal{N}(\v 1.5, 2 \m I)$. We then computed all pairwise feature distances, resulting in the histogram shown in Fig.~\figref{synthetic_distance_histogram}.
For the real experiment, we extracted both image-level features and patch-level features from all 209 normal (training) images and 63 anomaly (test) images of the \emph{bottle} sequence in MVTec AD using the pretrained DINO model from~\cite{caron2021emerging}.
Again, we calculated all pairwise feature distances over all pairs of patch-level features and over all pairs of image-level features, yielding  histograms in Figs.~\figref{real_distance_histogram} and \figref{real_patch_distance_histogram}.
While the histograms have different degrees of skewness, we observe the normal pairs are consistently the most likely to yield shorter pairwise distances compared to other types of pairs.

\section{Toy example of semantic anomaly detection}
\label{sec:toy_example_2}
We used CIFAR-10~\cite{krizhevsky2009learning} to conduct a toy experiment, setting the data to a scenario where the distribution of outliers is more spread out than the distribution of normals, consistent with our assumptions. 
The normal class is ``automobile'', and the outliers consist of the remaining classes in CIFAR-10. 
The contamination ratio (the ratio of outliers to normals) within the training dataset is set to 10\%. 
Unlike detecting patch-level defects, Local-Net in semantic anomaly detection outputs one anomaly score per image (because semantic anomalies are not divided into normal and abnormal regions within a single image).
For further details, please refer to Sec.~\ref{sec:semanti_anomaly_detection} for related results.

\begin{table*}[!t]
\renewcommand{\arraystretch}{1.2}
\fontsize{7}{8}\selectfont
    \centering
        \begin{tabular}{c|c|cccccc}
            \hline
            \multirow{2}{*}{Dataset} & \multirow{2}{*}{Method} & \multicolumn{6}{c}{Anomaly-to-normal ratio} \\ \cline{3-8} 
            & & $0\%$ & $1\%$ & $3\%$ & $5\%$ & $10\%$ & $20\%$ \\ \hline

            \multirow{4}{*}{MTD~\cite{magnetic}} & InReach~\cite{mcintosh2023inter} & \underline{83.55} / 72.02 & \underline{84.08} / 75.49 & 80.30 / 78.79 & 80.93 / 78.64 & 80.73 / 72.23 & 88.42 / 81.91 \\
            & SoftPatch~\cite{xi2022softpatch} & 76.11 / 90.63  & 77.19 / 91.53 & 79.61 / 92.67 & 80.51 / 91.48 &  83.61 / 87.52 & 85.26 / 93.49 \\
            & FUN-AD & 79.55 / \underline{94.75} & 82.51 / \textbf{94.58} & \underline{85.87} / \underline{94.97}  & \underline{85.35} / \underline{93.84} & \underline{85.12} / \underline{93.52} & \underline{94.74} / \underline{95.37} \\
            & FUN-AD* & \textbf{83.61} / \textbf{93.51} & \textbf{85.25} / \underline{93.59} & \textbf{87.52} / \textbf{95.22}  & \textbf{85.46} / \textbf{94.13} & \textbf{85.28} / \textbf{93.59} & \textbf{95.79} / \textbf{97.76} \\ \hline
        \end{tabular}
    \vspace{-2mm}
    \caption{Performance comparison of different fully-unsupervised anomaly detection methods across different anomaly-to-normal ratios on the contaminated MTD dataset (\textit{no overlap}). * indicates synthetic anomaly data has been utilized for training. The best results are in bold and the runner-ups are underlined.}
    \vspace{-4mm}    
    \label{tab:noise_ratio_v2}
\end{table*}

\section{Additional framework details}
\label{sec:additional_details}

\paragraph{Overall architecture}
FUN-AD comprises two sub-networks: a pretrained feature extractor $\mathcal{E}$ (to leverage semantic information, the self-supervised DINO\cite{caron2021emerging}) based on vision transformer (ViT)~\cite{dosovitskiy2020image} and the Local-Net model $\phi$ based on a simple multilayer perceptron (MLP) for detecting patch-level anomalies. 
$\mathcal{E}$ takes an image $I_i$ as input and outputs one class token and $P$ patch tokens.
To identify anomalies at the patch level, we concatenate the class token with each patch token to form a patch-level feature $\v f_{ij}\in\mathbb{R}^{D}$ for each image patch $X_{ij}$.
With abuse of notation, we represent $\ I_i = \{ X_{ij}\}_{j=1}^P$, where $P = HW / K^2$. In our setting, $H=W=224$, $K=8$, and thus $P=784$.
Since the class token is 768-dimensional and the patch token is 768-dimensional, $D=1536$. 
The local patch feature $\v f_{ij}$ serves as input to the Local-Net $\phi$, from which we obtain a normalized anomaly score using a sigmoid function.

\vspace{-3mm}
\paragraph{Model inference}
In the inference phase, FUN-AD performs anomaly detection and anomaly localization by predicting the anomaly score for a given input.
For anomaly detection, a test image $X_i$ is passed through $\mathcal{E}$ to extract the patch-level features $\{{\v f_{ij}}\}_{j=1}^P$. 
These features are passed through $\phi$ to obtain the patch-level anomaly scores. 
We then perform global max-pooling of these scores to calculate the image-level global anomaly score.
For anomaly localization, the patch-level anomaly scores are spatially arranged to form an anomaly score map, as shown in Fig. 5 in~\cite{authors24main}. 
As in\cite{xi2022softpatch, liu2023simplenet}, we then perform bilinear interpolation of the map with Gaussian smoothing ($\sigma=4$) to match the dimensions of the original image ($H\times W$).

\vspace{-3mm}
\paragraph{Implementation details}
The Local-Net has the FC[1536, 1024, 128, 1] structure. 
Leaky ReLU activation functions (slope: 0.2) are applied between layers, and the output layer uses the sigmoid function for outputting normalized anomaly score.
We use the RMSProp optimizer with momentum of 0.2 and learning rate of 2e-5 for training using the batch size of 32.
We set $\lambda=2.5$, $\tau_b=0.5$, $\tau_n=0.5$ and $\tau_c=0.9$ by default.
For each object/texture class, we train for 1500 epochs and choose the model with the best average of image-wise and pixel-wise AUROCs.
In Sec.~3 of the main paper~\cite{authors24main}, the real data consists of normal data in the training set and anomalous data in the test set.
Also, we set the patch size to 8, yielding one 768-dimensional image-level feature and $28^2$ 768-dimensional patch-level features for each normal or anomaly image.

\section{Additional ablation studies}
\paragraph{Effect of different contamination rates}
\label{sec:contamination}
Tab.~\ref{tab:noise_ratio_v2} demonstrates the performance of FUN-AD according to the contamination ratio in the training dataset.
Here, ``FUN-AD'' refers to the results obtained from training with the dataset without synthetic anomalies, while ``FUN-AD*'' refers to the results from training the FUN-AD framework with synthetic anomalies added at a rate of 5\% of the training dataset size.
Synthetic anomalies were created from a noisy (anomaly-present) dataset considering a fully unsupervised setting.
The results demonstrate that our proposed framework achieves state-of-the-art performance on texture-based dataset, highlighting its robustness across various types of anomalies.

\vspace{-3mm}
\paragraph{Inference time}
\label{sec:inferecne_time}
In an industrial setting, real-time anomaly detection is crucial. 
When comparing the inference speed with existing methods using the GPU RTX-4090 (refer to Tab.~\ref{tab:inference_time}), our method operates at an impressive speed of approximately 113 fps, outperforming other methods.

\vspace{-3mm}
\paragraph{Effect of weight on mutual smoothness loss}
\label{sec:mutual_smoothness_loss_ablation}
Tab.~\ref{tab:msl_weight_ablation} shows that optimal performance is achieved when $\lambda=2.5$ on MVTec AD.
In these results, $\lambda=0$ indicates that the pseudo-labeling method alone is sufficient for the network to learn from the normal and anomaly information and succeed in anomaly detection and localization. 
Additionally, when mutual-smoothness loss is applied, the anomaly detection performance improves with a weight value of $\lambda=2.5$ compared to using only pseudo-labeling.

\vspace{-3mm}
\paragraph{Effects of random sampling rate}
\label{sec:random_sampling}
Since Eq.7 needs to be calculated for pseudo-labeling, the training time overhead can be significant if computed with all feature vectors in the memory bank. 
However, applying corset sampling~\cite{sener2018active}, which has been used in a one-class classification environment, is difficult because we cannot assume that all the samples in the memory bank are normal. 
Therefore, we compare the performance by randomly sampling only a small percentage of the feature vectors in the memory bank.
Table~\ref{tab:random_ratio_ablation} shows the performance of anomaly detection and localization according to the sampling ratio. The performance does not vary significantly depending on the degree of sampling. 
This indicates that using a low sampling rate for efficient training does not result in significant performance degradation. 

\vspace{-3mm}
\paragraph{Effects of synthetic supervised loss}
The comparison with and without $y_\text{syn}$ in Tab.~\ref{tab:syn_sup} shows that our proposed pseudo labels perform better than those using Perlin masks to assign labels for synthetic anomalies. 
This indicates that our pseudo-labeling method is more effective for detecting real anomalies by identifying semantically anomalous regions and using them for training, rather than merely learning that regions with the Perlin noise are anomalous.

\vspace{-3mm}
\paragraph{Semantic anomaly detection}
\label{sec:semanti_anomaly_detection}
Detecting semantic anomalies also requires a fully unsupervised setting, and according to~\cite{wang2022hierarchical}, it is more similar to the real world when the training data is contaminated with abnormal samples. 
Therefore, we conducted experiments on the STR-10~\cite{wang2016unsupervised} and CIFAR-10~\cite{krizhevsky2009learning} datasets to verify the applicability of our framework.
We designated one class as the normal class and randomly sampled anomalies from the remaining classes to create contaminated unlabeled datasets with a 1:10 anomaly-to-normal ratio. 
The findings are presented in Tab.~\ref{tab:dataset_auroc}, demonstrating that although FUN-AD was originally developed for industrial anomaly detection, it effectively distinguishes between normal and abnormal classes under specific conditions.
In these experiments, where the normal class is singular and the abnormal classes encompass the remaining nine, the variation is substantial enough to validate the effectiveness of our assumptions and approach.

\begin{table}[!t]
\fontsize{7.5}{9}\selectfont
\centering
\renewcommand{\arraystretch}{1.1}
\begin{tabular}{c|c|c|c}
\hline
\textbf{Method}      & FUN-AD (\textit{Ours}) & SoftPatch~\cite{xi2022softpatch} & InReaCh~\cite{mcintosh2023inter} \\
\hline
\textbf{Throughput} (fps)    & \textbf{112.7} & 22.5           & 8.8          \\
\hline
\end{tabular}
\caption{
    Inference speeds achieved by different fully unsupervised anomaly detection algorithms on the VisA dataset.
}
\label{tab:inference_time}
\end{table}

\begin{table}[!t]
    \centering
    \resizebox{0.62\columnwidth}{!}{
        \begin{tabular}{c|cc}
            \toprule
            $\lambda$ & $\text{AUROC}_{\textit{\text{image}}}(\%)$ & $\text{AUROC}_{\textit{\text{pixel}}}(\%)$ \\ \midrule
            0    & 98.72 & 98.39 \\ 
            0.25 & 98.54 & 98.41 \\ 
            1 & 98.83 & 98.45 \\
            2.5 & \bf{98.95} & \bf{98.55} \\
            10 & 98.55 & 98.34 \\
            \bottomrule
        \end{tabular}
    }
    \vspace{-2mm}
    \caption{Ablation study of weights for the mutual smoothness loss. The optimal performance is achieved when $\lambda$ = 2.5 on MVTec AD.} 
    \label{tab:msl_weight_ablation}
\end{table}

\begin{table}[t]
    \centering
    \makebox[0pt][c]{\resizebox{0.8\columnwidth}{!}{
        \begin{tabular}{c|cc}
            \toprule
            Sampling ratio & $\text{AUROC}_{\textit{\text{image}}}(\%)$ & $\text{AUROC}_{\textit{\text{pixel}}}(\%)$ \\ 
            \midrule
            0.25 & 98.83 & {\bf 98.66} \\ 
            0.5 & 98.95 & 98.55 \\
            0.75 & \textbf{99.11} & 98.51 \\
            1.0 & 98.84 & 98.53 \\
            \bottomrule
        \end{tabular}
    }}
    \caption{Ablation study of the sampling rate when storing normally labeled feature vectors in memory banks. This demonstrates the capability of efficient training with a low sampling rate.}
    \label{tab:random_ratio_ablation}
\end{table}

\begin{table}[t]
    \centering
    \makebox[0pt][c]{\resizebox{0.8\columnwidth}{!}{
        \begin{tabular}{c|cc}
            \toprule
            Method & $\text{AUROC}_{\textit{\text{image}}}(\%)$ & $\text{AUROC}_{\textit{\text{pixel}}}(\%)$ \\ 
            \midrule
            w/o $y_\text{syn}$ & 97.83 & {\bf 97.51} \\ 
            w $y_\text{syn}$ & 98.95 & 98.55 \\
            \bottomrule
        \end{tabular}
    }}
    \caption{Ablation study of synthetic supervised loss. $y_\text{syn}$ indicates whether the mask of the synthetic anomaly is used or not.}
    \label{tab:syn_sup}
    \vspace{-4mm}
\end{table}

\begin{table}[t]
    \centering
    \resizebox{0.45\columnwidth}{!}{
        \begin{tabular}{c|c}
            \toprule
            Dataset  & AUROC (\%) \\ 
            \midrule
            CIFAR-10~\cite{krizhevsky2009learning} & 95.10 \\
            STL-10~\cite{wang2016unsupervised}   & 99.63 \\
            \bottomrule
        \end{tabular}
    }
    \caption{Average semantic anomaly detection results for scenarios (10, e.g., cat is normal) where each semantic class is normal.}
    \label{tab:dataset_auroc}
    \vspace{-4mm}
\end{table}

\begin{figure}[!t]
    \centering
    \subfloat[MVTec AD]{
        \centering
        \includegraphics[width=0.95\columnwidth]{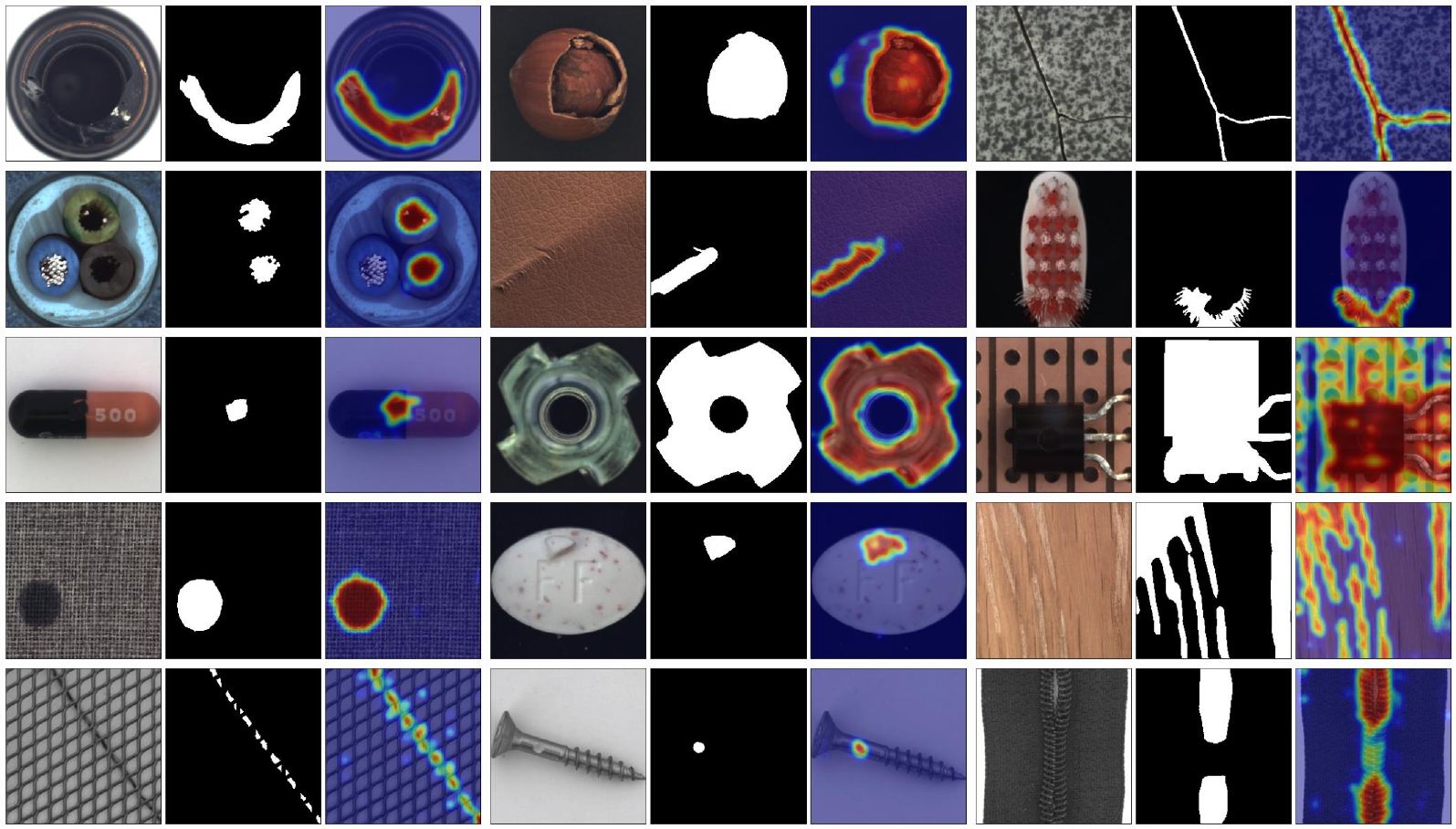}
        \label{fig:qualitative_results_mvtec}
    }
    \hfil
    \subfloat[VisA]{
        \centering
        \includegraphics[width=0.95\columnwidth]{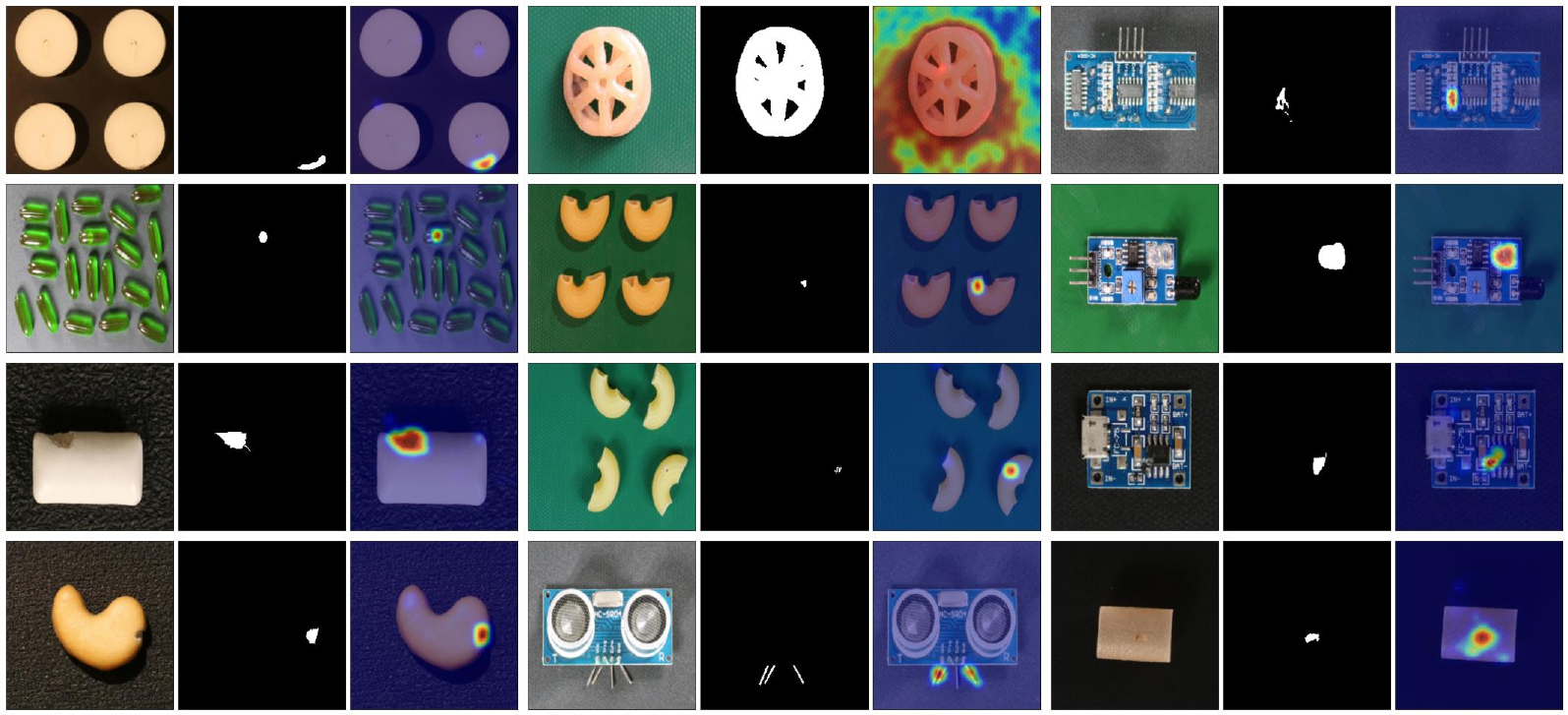}
        \label{fig:qualitative_results_VisA}
    }
    \caption{
         Visualization of anomaly detection results achieved by FUN-AD on the MVTec AD and VisA datasets. Each binary mask shows the anomaly segmentation map while each heatmap visualizes the anomaly region (red means likely to be an anomaly while blue means unlikely).
    }
\vspace{-4mm}
\label{fig:qualitative_results}
\end{figure}

\section{Qualitative results}
\label{sec:qualitative_results}
Fig.~\ref{fig:qualitative_results} shows some anomaly localization results yielded by FUN-AD. 
Each class is represented by three columns: the first column shows the RGB image, the second column shows the segmentation mask of the defect area, and the third column shows the anomaly score predicted by FUN-AD. 
Our method is not only effective at detecting large defects but also excels at clearly separating the boundary between normal and anomaly without ambiguity, even in the presence of very small defects. 
This is evident when comparing the ground-truth mask and the heatmap in Fig.~\ref{fig:qualitative_results}. 
These results demonstrate that FUN-AD is robust, particularly for very small defects, but not limited to large detects with higher confidence score compared to other models.

\section{Details of the experimental results}
\label{sec:detail_experiments}
We show the experimental results for all categories of \textit{overlap}, \textit{No overlap} for MVTec AD and VisA in Tab.~\ref{tab:detailed_noovelap_mvtec},~\ref{tab:detailed_ovelap_mvtec},~\ref{tab:detailed_noovelap_VisA},~\ref{tab:detailed_ovelap_VisA}. 
Each table presents image-wise AUROC (\%) / pixel-wise AUROC (\%), representing anomaly detection and localization performance, respectively.
The best results are in bold and the runner-ups are underlined. 
\textit{FUN-AD} places a stronger emphasis on local anomalies by utilizing Local-Net for inference. Consequently, it excels at detecting small defects in images with multiple instances, as observed in capsules and macaroni2 in VisA, outperforming other models in this regard.

\section{Limitations and broader impacts}
\paragraph{Limitations}
While FUN-AD is shown to work across many different unsupervised settings, it may be compromised if the feature diversity of the normal data is comparable to that of the anomalies, e.g. when one type of anomaly dominates.
Also, our analytic analysis in Sec.3.1 is limited to the case of normal and anomaly distributions following isotropic Gaussians.
Our approach still requires use of a pretrained feature extraction network such as DINO~\cite{caron2021emerging} for basic initialization.
Finally, FUN-AD yields suboptimal performance for scarce anomaly-to-normal ratios (0 to 1\%).

\vspace{-3mm}
\paragraph{Broader impacts}
Our approach can reduce the physical burden of human workers by reducing the manual labor required for annotating normal samples.
This allows reducing expenditure on data acquisition which in return may be invested towards improving the quality of product.

\begin{table*}[t]
\centering
\captionsetup{skip=0.1pt}

\footnotesize
\renewcommand{\arraystretch}{1.25}
\begin{tabular}{crrrrrrrr}
\hline
\multicolumn{1}{l}{Type} & \multicolumn{5}{c}{One-class classification} & \multicolumn{3}{c}{Fully unsupervised} \\ 
\cmidrule(lr){2-6} \cmidrule(lr){7-9}
Method & CS-Flow~\cite{rudolph2022fully} & PaDiM~\cite{defard2021padim} & PatchCore~\cite{roth2022towards} & SimpleNet~\cite{liu2023simplenet} & RealNet~\cite{zhang2024realnet} & SoftPatch~\cite{xi2022softpatch} & InReaCh~\cite{mcintosh2023inter} & FUN-AD (\textit{Ours}) \\

\hline
Bottle     & 98.7 / - & \underline{99.3} / 98.4 & \textbf{100.0} / 98.2 & \textbf{100.0} / 97.2 & \textbf{100.0} / 98.2 & \textbf{100.0} / \underline{98.7} & \textbf{100.0} / 98.3 & \textbf{100.0} / \textbf{99.2} \\
Cable      & 95.5 / - & 89.3 / 93.6 & 96.9 / 80.9 & 97.0 / 93.7 & 96.1 / 95.6 & \textbf{99.1} / \textbf{98.7} & 94.2 / \underline{97.5} & \underline{98.7} / 94.0 \\
Capsule    & 95.0 / - & 90.5 / 98.5 & 97.2 / 98.7 & 95.3 / 98.1 & \textbf{99.8} / \underline{98.6} & 95.8 / \textbf{98.9} & 49.7 / 93.4 & \underline{99.7} / 98.2 \\
Carpet     & 99.8 / - & \textbf{100.0} / 99.2 & 98.7 / 99.1 & \underline{99.4} / 98.8 & 98.6 / 97.7 & 99.1 / 99.4 & 98.4 / \underline{99.5} & \textbf{100.0} / \textbf{99.7} \\
Grid       & 96.5 / - & 93.4 / 95.6 & 96.2 / 98.9 & \underline{99.1} / 97.9 & \textbf{100.0} / \textbf{99.6} & 96.3 / 98.7 & 91.2 / 97.7 & \textbf{100.0} / \underline{99.4} \\
Hazelnut   & 95.1 / - & 92.7 / 98.0 & \textbf{100.0} / \underline{98.8} & 97.6 / 95.6 & \underline{99.9} / 98.5 & \textbf{100.0} / \underline{98.8} & 98.3 / 97.8 & \textbf{100.0} / \textbf{99.7} \\
Leather    & 98.6 / - & \textbf{100.0} / 99.3 & \textbf{100.0} / 99.2 & \textbf{100.0} / 98.9 & \textbf{100.0} / \underline{99.7} & \textbf{100.0} / 99.4 & \textbf{100.0} / 99.3 & \textbf{100.0} / \textbf{99.8} \\
Metal\_nut & 91.8 / - & 97.7 / 89.3 & 99.1 / 77.6 & 99.0 / 85.8 & \textbf{100.0} / 87.0 & \textbf{100.0} / 86.8 & 96.8 / \underline{95.1} & \textbf{100.0} / \textbf{99.5} \\
Pill       & 89.3 / - & 93.7 / 96.2 & 97.0 / 97.0 & 95.6 / 97.9 & \textbf{99.1} / \textbf{98.9} & 96.7 / 97.8 & 88.6 / 96.0 & \underline{98.5} / \underline{98.2} \\
Screw      & 79.3 / - & 84.5 / 98.4 & \underline{95.0} / \underline{98.6} & 89.0 / 97.8 & \textbf{98.8} / \textbf{99.4} & 94.5 / \textbf{99.4} & 79.7 / 98.3 & 94.8 / 98.1 \\
Tile       & 96.2 / - & 97.8 / 94.9 & 99.2 / 92.8 & 99.4 / 91.6 & \textbf{99.7} / \underline{97.8} & 98.7 / 96.3 & 99.0 / 97.3 & \underline{99.5} / \textbf{98.7} \\
Toothbrush & 92.7 / - & \textbf{100.0} / 98.8 & 99.7 / 98.8 & \textbf{100.0} / 98.4 & 98.3 / 95.0 & \underline{99.7} / 98.6 & 99.0 / \textbf{98.8} & 97.9 / \underline{98.7} \\
Transistor & 92.3 / - & 94.6 / 96.8 & 96.8 / 87.7 & 96.0 / 89.7 & 98.0 / 90.9 & \textbf{99.6} / 93.6 & 99.2 / \underline{97.1} & \underline{99.3} / \textbf{97.7} \\
Wood       & 90.5 / - & 98.1 / 94.3 & 96.6 / 95.6 & 99.7 / 92.9 & \underline{99.9} / \underline{97.3} & 98.1 / 95.1 & 95.3 / 92.3 & \textbf{100.0} / \textbf{98.0} \\
Zipper     & 95.1 / - & 88.2 / 98.4 & 98.7 / 98.1 & 98.7 / 95.2 & \underline{99.6} / \underline{98.9} & 97.5 / \underline{98.9} & 93.2 / 94.9 & \textbf{100.0} / \textbf{99.3} \\
\hline
Average & 93.8 / - & 94.7 / 96.6 & 98.1 / 94.7 & 97.7 / 95.3 & \textbf{99.2} / 96.9 & \underline{98.3} / \underline{97.3} & 92.2 / 96.9 & \textbf{99.2} / \textbf{98.6} \\
\hline
\end{tabular}
\vspace{3mm}
\caption{
    Detailed results for MVTec AD in the \textit{No overlap} setting.
}
\label{tab:detailed_noovelap_mvtec}
\end{table*}

\begin{table*}[t]
\centering
\captionsetup{skip=1pt}
\footnotesize
\renewcommand{\arraystretch}{1.25}
\begin{tabular}{crrrrrrrr}
\hline
\multicolumn{1}{l}{Type} & \multicolumn{5}{c}{One-class classification} & \multicolumn{3}{c}{Fully unsupervised} \\ 
\cmidrule(lr){2-6} \cmidrule(lr){7-9}
Method & CS-Flow~\cite{rudolph2022fully} & PaDiM~\cite{defard2021padim} & PatchCore~\cite{roth2022towards} & SimpleNet~\cite{liu2023simplenet} & RealNet~\cite{zhang2024realnet} & SoftPatch~\cite{xi2022softpatch} & InReaCh~\cite{mcintosh2023inter} & FUN-AD (\textit{Ours}) \\

\hline
Bottle     & 67.4 / - & 79.9 / 96.0 & 89.4 / 68.2 & 76.2 / 46.3 & \underline{99.3} / 97.7 & \textbf{100.0} / 93.1 & \textbf{100.0} / \underline{98.4} & \textbf{100.0} / \textbf{99.2} \\
Cable      & 72.7 / - & 68.0 / 86.0 & 87.1 / 65.0 & 41.2 / 51.7 & 88.3 / 93.5 & \textbf{99.3} / \textbf{98.4} & 94.6 / \underline{97.8} & \underline{96.9} / 95.4 \\
Capsule    & 77.7 / - & 81.4 / 98.3 & 88.8 / 92.3 & 43.0 / 61.0 & \textbf{97.8} / \textbf{99.2} & 95.3 / \underline{98.7} & 52.1 / 94.3 & \underline{97.5} / 98.1 \\
Carpet     & 68.4 / - & 89.0 / 97.1 & 75.4 / 64.6 & 67.6 / 71.1 & 98.9 / 98.2 & \underline{99.7} / 98.5 & 98.7 / \underline{99.4} & \textbf{100.0} / \textbf{99.7} \\
Grid       & 52.5 / - & 52.0 / 85.7 & 61.3 / 60.3 & 57.6 / 45.4 & \textbf{100.0} / \underline{99.1} & 97.1 / 97.6 & 92.4 / 97.6 & \underline{99.8} / \textbf{99.2} \\
Hazelnut   & 42.1 / - & 43.2 / 95.8 & 67.0 / 59.7 & 69.4 / 64.1 & \underline{99.6} / \underline{98.5} & \textbf{100.0} / 95.4 & 98.7 / 97.6 & \textbf{100.0} / \textbf{99.7} \\
Leather    & 72.9 / - & \underline{94.7} / 97.2 & 81.0 / 66.9 & 45.8 / 31.5 & \textbf{100.0} / \underline{99.4} & \textbf{100.0}/ 99.3 & \textbf{100.0} / 99.0 & \textbf{100.0} / \textbf{99.8} \\
Metal\_nut & 70.1 / - & 74.6 / 79.9 & 90.2 / 62.7 & 60.9 / 60.7 & 97.2 / 86.3 & \underline{99.7} / 77.0 & 97.2 / \underline{98.4} & \textbf{99.8} / \textbf{99.4} \\
Pill       & 72.8 / - & 76.5 / 95.9 & 84.5 / 90.8 & 49.1 / 55.6 & \underline{96.5} / \textbf{98.5} & 94.6 / \underline{97.3} & 89.1 / 96.1 & \textbf{98.1} / \textbf{98.5} \\
Screw      & 58.0 / - & 62.5 / 96.7 & 78.0 / 77.2 & 53.6 / 61.9 & 91.2 / \textbf{98.9} & \textbf{93.5} / 94.5 & 79.9 / 98.3 & \underline{93.4} / \underline{98.4} \\
Tile       & 69.8 / - & 71.0 / 72.6 & 86.8 / 63.3 & 66.9 / 33.0 & 98.6 / 94.2 & \textbf{99.4} / 94.0 & \underline{99.1} / \underline{97.7} & 99.0 / \textbf{98.8} \\
Toothbrush & 83.9 / - & 80.0 / 95.3 & 95.3 / 91.1 & 50.6 / 34.9 & \underline{99.2} / 94.1 & \textbf{99.7} / \textbf{98.8} & 98.3 / \textbf{98.8} & 98.6 / \textbf{98.8} \\
Transistor & 43.8 / - & 45.8 / 90.8 & 72.1 / 61.4 & 61.5 / 48.3 & 88.5 / 87.6 & \textbf{99.1} / 90.4 & \underline{99.0} / \underline{95.2} & 98.7 / \textbf{97.7} \\
Wood       & 54.3 / - & 66.8 / 90.2 & 76.0 / 62.0 & 79.2 / 64.6 & 97.6 / \underline{96.1} & \underline{98.9} / 95.2 & 95.2 / 91.8 & \textbf{100.0} / \textbf{98.0} \\
Zipper     & 75.9 / - & 77.3 / 96.9 & 90.7 / 76.4 & 66.9 / 63.4 & \underline{99.7} / 98.5 & 97.6 / \underline{98.7} & 91.6 / 94.4 & \textbf{99.9} / \textbf{99.2} \\
\hline
Average & 65.5 / - & 70.8 / 91.6 & 81.6 / 70.8 & 59.3 / 52.9 & 96.8 / 96.0 & \underline{98.3} / 95.1 & 92.4 / \underline{97.2} & \textbf{98.8} / \textbf{98.6} \\
\hline
\end{tabular}
\vspace{3mm}
\caption{
    Detailed results for MVTec AD in the \textit{Overlap} setting.
}
\label{tab:detailed_ovelap_mvtec}
\end{table*}

\clearpage

\begin{table*}[t]
\centering
\captionsetup{skip=1pt}
\footnotesize
\renewcommand{\arraystretch}{1.25}
\begin{tabular}{crrrrrrrr}
\hline
\multicolumn{1}{l}{Type} & \multicolumn{5}{c}{One-class classification} & \multicolumn{3}{c}{Fully unsupervised} \\ 
\cmidrule(lr){2-6} \cmidrule(lr){7-9}
Method & CS-Flow~\cite{rudolph2022fully} & PaDiM~\cite{defard2021padim} & PatchCore~\cite{roth2022towards} & SimpleNet~\cite{liu2023simplenet} & RealNet~\cite{zhang2024realnet} & SoftPatch~\cite{xi2022softpatch} & InReaCh~\cite{mcintosh2023inter} & FUN-AD (\textit{Ours}) \\

\hline
Candle     & 86.7 / - & 91.2 / 99.4 & 95.0 / 99.2 & 93.9 / 97.9 & \textbf{94.7} / \textbf{99.6} & 93.8 / \textbf{99.6} & 90.1 / 98.7 & \underline{94.4} / \underline{99.5} \\
Capsules   & 68.2 / - & 56.6 / 94.0 & 69.5 / 96.2 & 76.5 / 96.8 & \underline{82.5} / \underline{98.9} & 69.2 / 97.7 & 57.9 / 93.0 & \textbf{93.8} / \textbf{99.5} \\
Cashew     & 86.1 / - & 90.0 / 99.0 & 94.7 / 99.0 & 90.7 / 99.5 & 84.3 / 98.2 & \underline{95.2} / \underline{99.1} & 77.1 / 98.2 & \textbf{96.6} / \textbf{99.7} \\
Chewinggum & 93.8 / - & 95.2 / 98.9 & 98.4 / 98.5 & 96.0 / 97.5 & \textbf{99.8} / \textbf{99.9} & 98.7 / 99.1 & 77.8 / 98.2 & \underline{99.3} / \underline{99.7} \\
Fryum      & 78.0 / - & 89.1 / \underline{97.7} & 89.4 / 91.2 & 91.5 / 94.8 & 89.1 / 94.5 & \underline{92.0} / 96.0 & 86.3 / 96.4 & \textbf{96.9} / \textbf{98.2} \\
Macaroni1  & 81.4 / - & 83.2 / \underline{99.1} & 86.5 / 97.3 & 88.7 / 98.1 & \textbf{98.1} / \textbf{99.9} & 89.8 / 98.8 & 83.7 / 98.0 & \underline{96.4} / \textbf{99.9} \\
Macaroni2  & 60.6 / - & 60.4 / 95.6 & 64.8 / 89.3 & 72.1 / 94.6 & \textbf{90.0} / \textbf{99.6} & 57.6 / 95.1 & 57.4 / 95.8 & \underline{87.0} / \underline{99.1} \\
PCB1       & 90.7 / - & 94.2 / \underline{99.6} & 93.0 / 86.5 & 93.0 / 94.8 & 93.8 / 99.4 & \underline{95.1} / \textbf{99.8} & 93.8 / \underline{99.6} & \textbf{96.2} / \underline{99.6} \\
PCB2       & 85.8 / - & 91.8 / 99.2 & \textbf{96.3} / 98.8 & 94.8 / \underline{99.0} & \underline{95.5} / 96.9 & 93.9 / \textbf{99.3} & 91.9 / 98.7 & 91.2 / 97.8 \\
PCB3       & 84.3 / - & 85.7 / 99.1 & 93.8 / 96.9 & \underline{95.2} / 99.2 & \textbf{95.9} / 99.1 & 92.3 / \textbf{99.4} & 93.3 / \underline{99.3} & 91.4 / 98.6 \\
PCB4       & 95.3 / - & 97.1 / 98.2 & 98.0 / 96.3 & 97.8 / 96.1 & 98.9 / 98.9 & \underline{99.2} / \underline{99.1} & \textbf{99.7} / \textbf{99.5} & 97.6 / \textbf{99.5} \\
Pipe\_fryum& 77.3 / - & 95.2 / 98.3 & 98.5 / \underline{99.2} & 94.6 / \underline{99.7} & 98.8 / 99.3 & 98.8 / 99.5 & 96.7 / \textbf{99.8} & \textbf{99.3} / \textbf{99.8} \\
\hline
Average & 82.3 / - & 85.8 / 98.3 & 89.8 / 95.7 & 90.4 / 96.7 & \underline{93.5} / \underline{98.7} & 89.6 / 98.5 & 83.8 / 97.6 & \textbf{95.0} / \textbf{99.2} \\
\hline
\end{tabular}
\vspace{3mm}
\caption{
    Detailed results for VisA in the \textit{No overlap} setting.
}
\label{tab:detailed_noovelap_VisA}
\end{table*}

\begin{table*}[!t]
\centering
\captionsetup{skip=1pt}
\footnotesize
\renewcommand{\arraystretch}{1.25}
\begin{tabular}{crrrrrrrr}
\hline
\multicolumn{1}{l}{Type} & \multicolumn{5}{c}{One-class classification} & \multicolumn{3}{c}{Fully unsupervised} \\ 
\cmidrule(lr){2-6} \cmidrule(lr){7-9}
Method & CS-Flow~\cite{rudolph2022fully} & PaDiM~\cite{defard2021padim} & PatchCore~\cite{roth2022towards} & SimpleNet~\cite{liu2023simplenet} & RealNet~\cite{zhang2024realnet} & SoftPatch~\cite{xi2022softpatch} & InReaCh~\cite{mcintosh2023inter} & FUN-AD (\textit{Ours}) \\

\hline
Candle     & 68.1 / - & 79.7 / \underline{99.1} & 85.3 / 87.5 & 47.6 / 49.0 & \textbf{95.3} / \textbf{99.4} & 93.7 / \textbf{99.4} & 85.2 / 95.5 & \underline{93.8} / \textbf{99.4} \\
Capsules   & 48.6 / - & 45.3 / 78.9 & 65.2 / 83.9 & 50.3 / 50.2 & \underline{82.6} / \underline{96.8} & 70.5 / 89.2 & 53.1 / 89.0 & \textbf{93.3} / \textbf{99.4} \\
Cashew     & 64.4 / - & 72.0 / 85.2 & 86.8 / 85.7 & 61.4 / 60.8 & 88.9 / 96.6 & \underline{94.1} / \underline{98.9} & 75.5 / 91.0 & \textbf{96.9} / \textbf{99.7} \\
Chewinggum & 57.8 / - & 76.2 / 82.1 & 87.9 / 78.1 & 51.9 / 59.9 & \textbf{99.9} / \underline{99.4} & 97.9 / 98.9 & 76.1 / 90.4 & \underline{99.2} / \textbf{99.7} \\
Fryum      & 73.1 / - & 71.3 / 92.9 & 84.3 / 82.5 & 51.9 / 80.0 & 86.1 / \underline{94.8} & \underline{92.9} / 92.4 & 78.1 / 94.7 & \textbf{96.5} / \textbf{98.2} \\
Macaroni1  & 65.1 / - & 68.8 / 97.5 & 80.2 / 84.0 & 45.7 / 49.3 & \textbf{98.1} / \textbf{99.8} & \underline{89.0} / 97.2 & 75.5 / 88.1 & \underline{96.7} / \textbf{99.8} \\
Macaroni2  & 48.2 / - & 48.4 / 89.7 & 58.2 / 80.4 & 51.2 / 57.0 & \textbf{88.8} / \textbf{99.2} & 56.5 / 85.9 & 51.0 / \underline{91.4} & \underline{87.7} / \textbf{99.2} \\
PCB1       & 72.5 / - & 77.4 / 98.0 & 87.3 / 71.4 & 44.7 / 60.8 & 93.2 / \underline{98.9} & \underline{95.9} / \textbf{99.6} & 94.4 / 97.2 & \textbf{96.7} / \textbf{99.6} \\
PCB2       & 68.7 / - & 74.1 / 96.4 & 86.3 / 83.5 & 43.9 / 46.8 & 91.0 / \underline{96.5} & \textbf{93.0} / \textbf{98.0} & 86.7 / 94.1 & \underline{91.5} / \textbf{98.0} \\
PCB3       & 67.5 / - & 69.6 / 97.0 & 85.6 / 76.6 & 53.7 / 65.8 & 90.0 / 96.3 & \textbf{92.4} / \textbf{98.7} & 83.1 / 97.3 & \underline{91.7} / \underline{98.2} \\
PCB4       & 76.3 / - & 82.1 / 95.7 & 93.9 / 83.0 & 46.9 / 68.0 & 97.0 / 96.6 & \textbf{99.2} / \underline{97.9} & \underline{99.0} / 96.7 & 97.8 / \textbf{99.5} \\
Pipe\_fryum& 61.8 / - & 76.2 / 95.7 & 87.1 / 82.6 & 56.2 / 47.2 & 98.1 / 97.7 & \underline{98.4} / \underline{99.4} & 84.4 / 97.5 & \textbf{99.4} / \textbf{99.8} \\
\hline
Average & 64.3 / - & 70.1 / 92.3 & 82.3 / 81.6 & 50.4 / 57.9 & \underline{92.4} / \underline{97.7} & 89.5 / 96.3 & 78.5 / 93.6 & \textbf{95.1} / \textbf{99.2} \\
\hline
\end{tabular}
\vspace{3mm}
\caption{
    Detailed results for VisA in the \textit{Overlap} setting.
}
\label{tab:detailed_ovelap_VisA}
\end{table*}
\end{document}